\documentclass[sigconf, 10pt]{acmart}
\newcommand{\ignore}[1]{}
\usepackage{fancyhdr}
\usepackage[normalem]{ulem}
\usepackage[ruled,vlined,linesnumbered]{algorithm2e}
\usepackage{graphicx}
\usepackage{textcomp}
\usepackage{xcolor}
\usepackage{multirow}
\usepackage{amsmath,amssymb,amsfonts}

\copyrightyear{2019} 
\acmYear{2019} 
\setcopyright{acmcopyright}
\acmConference[ISCA '19]{The 46th Annual International Symposium on Computer Architecture}{June 22--26, 2019}{Phoenix, AZ, USA}
\acmBooktitle{The 46th Annual International Symposium on Computer Architecture (ISCA '19), June 22--26, 2019, Phoenix, AZ, USA}
\acmPrice{15.00}
\acmDOI{10.1145/3307650.3322270}
\acmISBN{978-1-4503-6669-4/19/06}

\begin{document}

\title{A Stochastic-Computing based Deep Learning Framework using Adiabatic Quantum-Flux-Parametron Superconducting Technology}

\author{Ruizhe Cai}
\author{Ao Ren}
    \affiliation{%
        \institution{Northeastern University}
        \state{USA} 
    }
    \email{{cai.ruiz, ren.ao}@husky.neu.edu}
    
\author{Olivia Chen}
    \affiliation{%
        \institution{Yokohama National University}
        \state{Japan} 
    }
    \email{chen-olivia-pg@ynu.ac.jp}
    
\author{Ning Liu}
\author{Caiwen Ding}
    \affiliation{%
        \institution{Northeastern University}
        \state{USA} 
    }
    \email{{liu.ning, ding.ca}@husky.neu.edu}

\author{Xuehai Qian}
    \affiliation{%
        \institution{University of Southern California}
        \state{USA}
    }
    \email{xuehai.qian@usc.edu}

\author{Jie Han}
    \affiliation{%
        \institution{University of Alberta}
        \state{Canada}
    }
    \email{jhan8@ualberta.ca}

\author{Wenhui Luo}
    \affiliation{%
        \institution{Yokohama National University}
        \state{Japan} 
    }
    \email{luo-wenhui-xn@ynu.ac.jp}
    
\author{Nobuyuki Yoshikawa}
    \affiliation{%
        \institution{Yokohama National University}
        \state{Japan} 
    }
    \email{nyoshi@ynu.ac.jp}
    
\author{Yanzhi Wang}
    \affiliation{%
        \institution{Northeastern University}
        \state{USA} 
    }
    \email{yanz.wang@northeastern.edu}

\renewcommand{\shortauthors}{R. Cai and A. Ren et al.}
\renewcommand{\shorttitle}{A SC based Deep Learning Framework using AQFP Superconducting Technology}

\begin{abstract}
The \emph{Adiabatic Quantum-Flux-Parametron} (AQFP) superconducting technology has been recently developed, which achieves the highest energy efficiency among superconducting logic families, potentially $10^4$-$10^5$ gain compared with state-of-the-art CMOS. In 2016, the successful fabrication and testing of AQFP-based circuits with the scale of 83,000 JJs have demonstrated the scalability and potential of implementing large-scale systems using AQFP. As a result, it will be promising for AQFP in high-performance computing and deep space applications, with Deep Neural Network (DNN) inference acceleration as an important example. 

Besides ultra-high energy efficiency, AQFP exhibits two unique characteristics: the deep pipelining nature since each AQFP logic gate is connected with an AC clock signal, which increases the difficulty to avoid RAW hazards; the second is the unique opportunity of true random number generation (RNG) using a single AQFP buffer, far more efficient than RNG in CMOS. We point out that these two characteristics make AQFP especially compatible with the \emph{stochastic computing} (SC) technique, which uses a time-independent bit sequence for value representation, and is compatible with the deep pipelining nature. Further, the application of SC has been investigated in DNNs in prior work, and the suitability has been illustrated as SC is more compatible with approximate computations.

This work is the first to develop an SC-based DNN acceleration framework using AQFP technology. 
The deep-pipelining nature of AQFP circuits translates into the difficulty in designing accumulators/counters in AQFP, which makes the prior design in SC-based DNN not suitable. We overcome this limitation taking into account different properties in CONV and FC layers: (i) the inner product calculation in FC layers has more number of inputs than that in CONV layers; (ii) accurate activation function is critical in CONV rather than FC layers. Based on these observations, we propose (i) accurate integration of summation and activation function in CONV layers using bitonic sorting network and feedback loop, and (ii) low-complexity categorization block for FC layers based on chain of majority gates. For complete design we also develop (i) ultra-efficient stochastic number generator in AQFP, (ii) a high-accuracy sub-sampling (pooling) block in AQFP, and (iii) majority synthesis for further performance improvement and automatic buffer/splitter insertion for requirement of AQFP circuits. Experimental results suggest that the proposed SC-based DNN using AQFP can achieve up to $6.8\times 10^4$ times higher energy efficiency compared to CMOS-based implementation while maintaining $96\%$ accuracy on the MNIST dataset.

\end{abstract}

\maketitle
\section{Introduction}
Wide-ranging applications of deep neural networks (DNNs) in image classification, computer vision, autonomous driving, embedded and IoT systems, etc., call for high-performance and energy-efficient implementation of the inference phase of DNNs. 
To simultaneously achieve high performance and energy efficiency, \emph{hardware acceleration of DNNs}, including  FPGA- and ASIC-based implementations, has been extensively investigated  \cite{kwon2018maeri,chen2014diannao,judd2016stripes,sharma2016high,chen2014dadiannao,venkataramani2017scaledeep,reagen2016minerva,han2016eie,du2015shidiannao,song2018situ,mahajan2016tabla,han2017ese,umuroglu2017finn,zhao2017accelerating,suda2016throughput,qiu2016going,chen2017eyeriss,moons201714,desoli201714,whatmough201714,bang201714,sim201614,zhang2016caffeine,zhang2016energy,Cai:2018:VHA:3296957.3173212}. 
However, most of these designs are CMOS based, and suffer from a performance limitation because the Moore's Law is reaching its end. 

Being widely-known for low energy dissipation and ultra-fast switching speed, Josephson Junction (JJ) based superconductor logic families have been proposed and implemented to process analog and digital signals \cite{80745}. It has been perceived to be an important candidate to replace state-of-the-art CMOS due to the superior potential in operation speed and energy efficiency, as recognized by the U.S. IARPA C3 and SuperTools Programs and Japan MEXT-JSPS Project.
Adiabatic quantum-flux-parametron (AQFP) logic is an energy-efficient superconductor logic family based on the quantum-flux-parametron (QFP)\cite{Goto}. AQFP logic achieves high energy efficiency by adopting adiabatic switching \cite{Likharev}, in which the potential energy profile evolves from a single well to a double well so that the logic state can change quasi-statically.  The energy-delay-product (EDP) of the AQFP circuits fabricated using processes such as the AIST standard process 2 (STP2) \cite{nagasawa1995380} and the MIT-LL SFQ process \cite{tolpygo2016advanced}, is only three orders of magnitude larger than the quantum limit \cite{Takeuchi2014QuantumLimits}. It can potentially achieve $10^4$-$10^5$ energy efficiency gain compared with state-of-the-art CMOS (even two order of magnitude energy efficiency gain when cooling energy is accounted for), with a clock frequency of several GHz. Recently, the successful fabrication and testing of AQFP-based implementations with the scale of 83,000 JJs, have demonstrated the scalability and potential of implementing large-scale circuits/systems using AQFP \cite{83k}. As a result, it will be promising for the AQFP technology in high-performance computing, with DNN inference acceleration an important example. 
\begin{figure}[t]
\centering
\includegraphics[width=0.18\textwidth,angle =90]{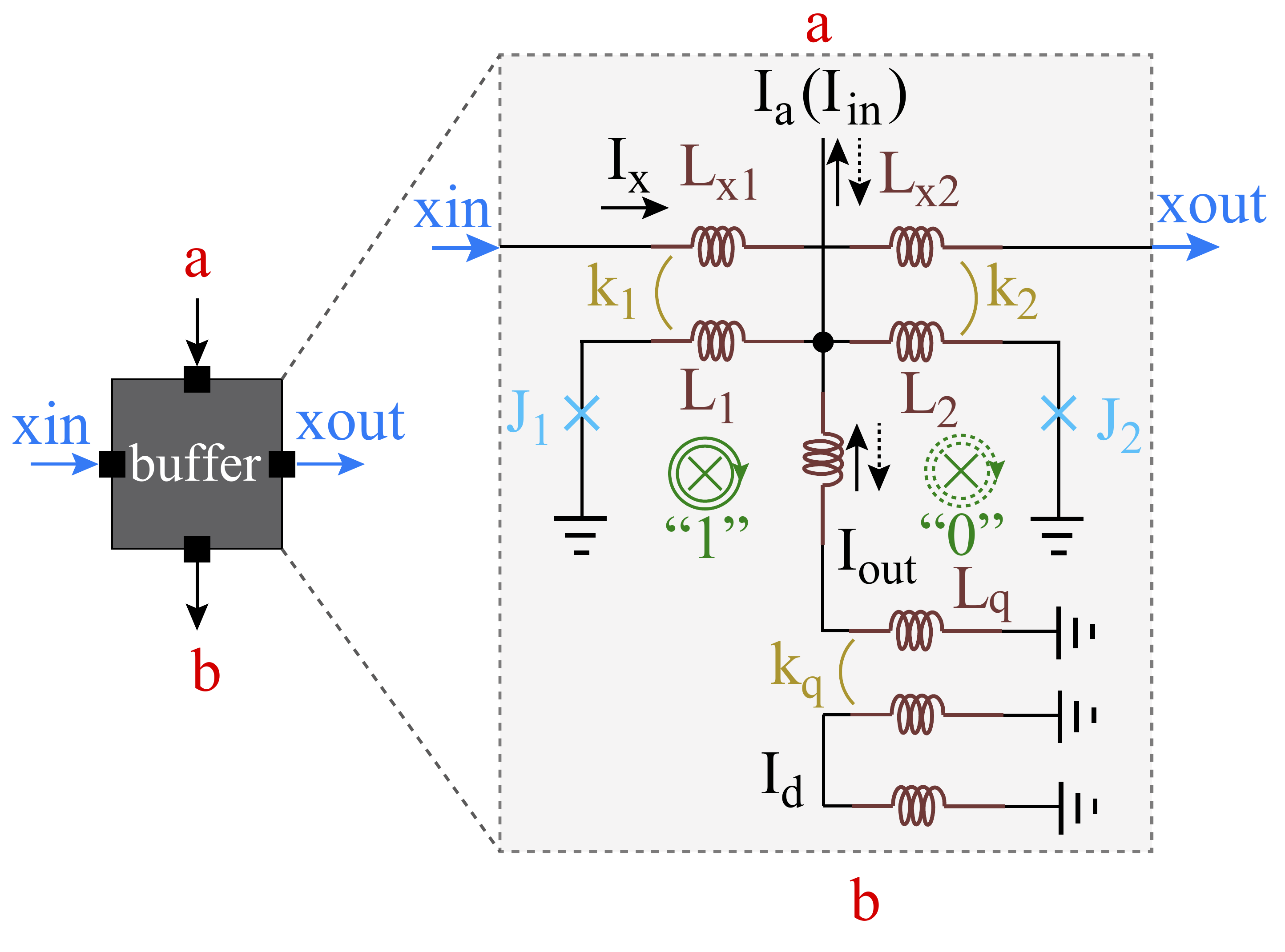}
\caption{Junction level schematic of an AQFP buffer.}
\label{fig:aqfp_buffer}
\end{figure}

The AQFP technology uses AC bias/excitation currents as both multi-phase clock signal and power supply \cite{takeuchi2013adiabatic} to mitigate the power consumption overhead of DC bias in other superconducting logic technologies. Besides the ultra-high energy efficiency, AQFP exhibits two unique characteristics. The first is the deep pipelining nature since each AQFP logic gate is connected with an AC clock signal and occupies one clock phase, which increases the difficulty to avoid RAW (Read after Write) hazards in conventional binary computing. The second is the unique opportunity of true random number generation (RNG) using a single AQFP buffer (double JJ) in AQFP \cite{takeuchi2013measurement,gentle2006random}, which is far more efficient than RNG in CMOS. 

We point out that these two characteristics make AQFP technology especially compatible with the \emph{stochastic computing} (SC) technique \cite{gaines1967stochastic}\cite{alaghi2013survey}, which allows the implementation of 64 basic operations using extremely small hardware footprint. SC uses a time-independent bit sequence for value representation, and lacks RAW dependency among the bits in the stream. As a result it is compatible with the deep-pipelining nature of superconducting logic. Furthermore, one important limiting factor of SC, i.e., the overhead of RNG \cite{l2012random}\cite{niederreiter1992random}, can be largely mitigated in AQFP.

SC is inherently an approximate computing technique \cite{gaines1969stochastic}\cite{xiu2010numerical}, and there have been disputes about the suitability of SC for precise computing applications \cite{ardakani2017vlsi}. On the other hand, the DNN inference engine is essentially an approximate computing application. This is because the final classification result depends on the relative score/logit values of different classes, instead of absolute values. Recent work \cite{ren2017sc}\cite{ardakani2017vlsi} have pointed out the suitability of SC for DNN acceleration, and \cite{wang2018universal} has further proved the equivalence between SC-based DNN and \emph{binary neural networks}, where the latter originate from deep learning society \cite{courbariaux2016binarized}. All the above discussions suggest the potential to build SC-based DNN acceleration using AQFP technology.

This paper is the first to develop an SC-based DNN acceleration framework using AQFP superconducting technology. We adopt bipolar format in SC because weights and inputs can be positive or negative, and build stochastic number generation block in AQFP with ultra-high efficiency. The deep-pipelining nature of AQFP circuits translates into the difficulty in designing accumulators/counters in AQFP, which makes the prior design in SC-based DNN \cite{ren2017sc} not suitable. We overcome this limitation taking into account different properties in CONV and FC layers.In summary, the contributions are listed as follows: (i) ultra-efficient stochastic number generator in AQFP; (ii) integration of summation and activation function using bitonic sorting in CONV layers; (iii) a high-accuracy sub-sampling block in AQFP; (iv) low-complexity categorization block for FC layers; (v) majority synthesis for further performance improvement and automatic buffer/splitter insertion for AQFP requirement.
Experimental results suggest that the proposed SC-based DNN using AQFP can achieve up to $6.9\times 10^4$ higher energy efficiency compared to its CMOS implemented counterpart and $96\%$ accuracy on the MNIST dataset \cite{lecun-mnisthandwrittendigit-2010}. The proposed blocks can be up to $7.76\times 10^5$ times more energy efficient than CMOS implementation. Simulation results suggest that the proposed sorter-based DNN blocks can achieve extremely low inaccuracy. In addition, we have successfully verified the functionality of a feature extraction chip, which is fabricated using the AIST $10kA/cm^2$ Niobium high-speed standard process (HSTP), embedded in a cryoprobe and inserted into a liquid Helium Dewar to cool down to 4.2K.

\section{Background}\label{sec:background}

\subsection{AQFP Superconducting Logic}
\label{sec:background_aqfp}
As shown in Fig. \ref{fig:aqfp_buffer}, the basic logic structure of AQFP circuits is buffers consisting of a double-Josephson-Junction SQUID \cite{clarke2006squid}. An AQFP logic gate is mainly driven by AC-power, which serves as both excitation current and power supply. By applying excitation current $I_x$, typically in the order of hundreds of micro-amperes, excitation fluxes are applied to the superconducting loops via inductors $L_1$, $L_2$, $L_{x1}$ and $L_{x2}$. Depending on the small input current $I_{in}$, either the left or the right loop stores one single flux quantum. Consequently, the device is capable of acting as a \emph{buffer cell} as the logic state can be represented by the direction of the output current $I_{out}$. The storage position of the quantum flux in the left or the right loop can be encoded as logic `1' or `0'.

\begin{figure}[t]
\centering
\includegraphics[width=0.4\textwidth]{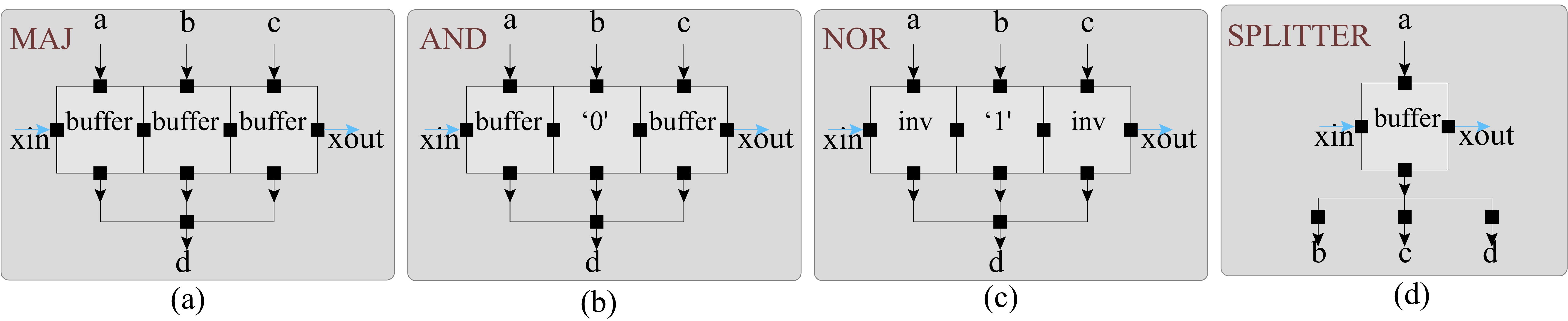}
\caption{Example of AQFP logic gates.}
\label{fig:aqfp_ex}
\end{figure}

The AQFP inverter and constant cell are designed from AQFP buffer. The AQFP inverter is implemented by negating the coupling coefficient of the output transformer in the AQFP buffer, while the constant gate in AQFP is designed using asymmetry excitation flux inductance in the AQFP buffer. The AQFP splitter is also implemented based on the AQFP buffer as shown in Fig. \ref{fig:aqfp_ex}. Please note that, unlike CMOS gates, which are connected to fan-out directly, all AQFP gates need to be connected to splitters for fan-out. 

The AQFP standard cell library is built via the \textit{minimalist design approach} \cite{takeuchi2015adiabatic}, in other words, designing more complicated gates using a bottom-up manner. Logic gate shown in Fig. \ref{fig:aqfp_ex} (a) is a majority gate as the output, `d', depending on the number of `1's at inputs, `a' to `c'. Replacing any buffer with inverter can produce a variation of the majority gate (such as the minority gate). As illustrated in Fig. \ref{fig:aqfp_ex} (b)-(c), an AQFP AND gate is implemented by two buffers and one constant `0' while an AQFP NOR gate is implemented by two inverters and one constant `1'. 
\begin{figure}[t]
\centering
\includegraphics[width=0.35\textwidth]{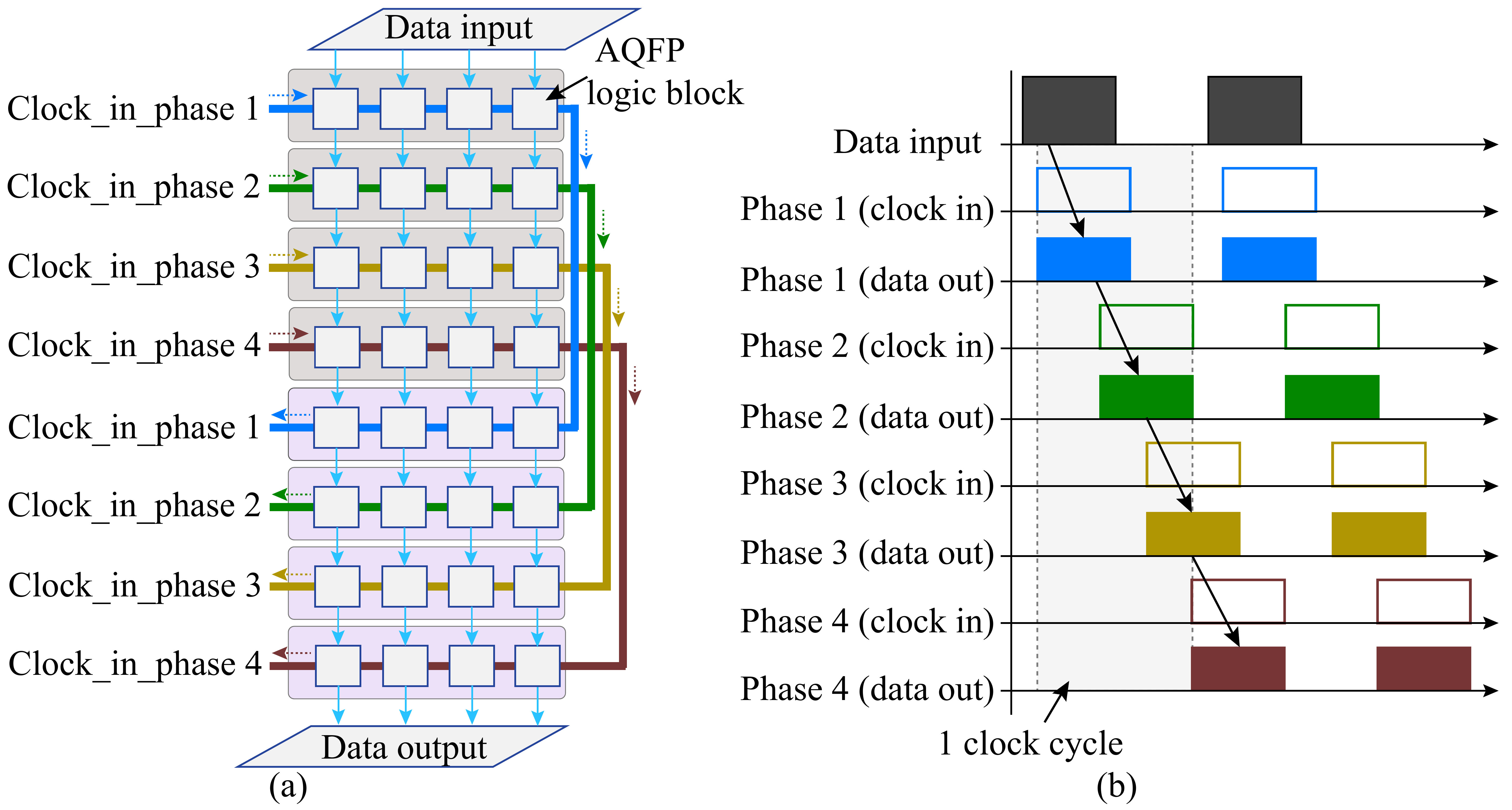}
\caption{(a). Four phase clocking scheme for AQFP circuits; (b). Data propagation between neighbouring clock phases.}
\label{fig:aqfp_flow}
\end{figure}
Unlike the conventional CMOS technology, both combinational and sequential AQFP logic cells are driven by AC-power. In addition, the AC power serves as clock signal to synchronize the outputs of all gates in the same clock phase. Consequently, data propagation in AQFP circuits requires overlapping of clock signals from neighbouring phases. Fig. \ref{fig:aqfp_flow} presents an example of typical clocking scheme of AQFP circuits and data flow between neighbouring clock phases. In this clocking scheme, each AQFP logic gate is connected with an AC clock signal and occupies one clock phase, which makes AQFP circuits "deep-pipelining" in nature. Such clock-based synchronization characteristic also requires that all inputs for each gate should have the same delay (clock phases) from the primary inputs.


\subsection{Stochastic Computing}
\begin{figure}[t]
\begin{center}
\includegraphics[width = 0.8\linewidth]{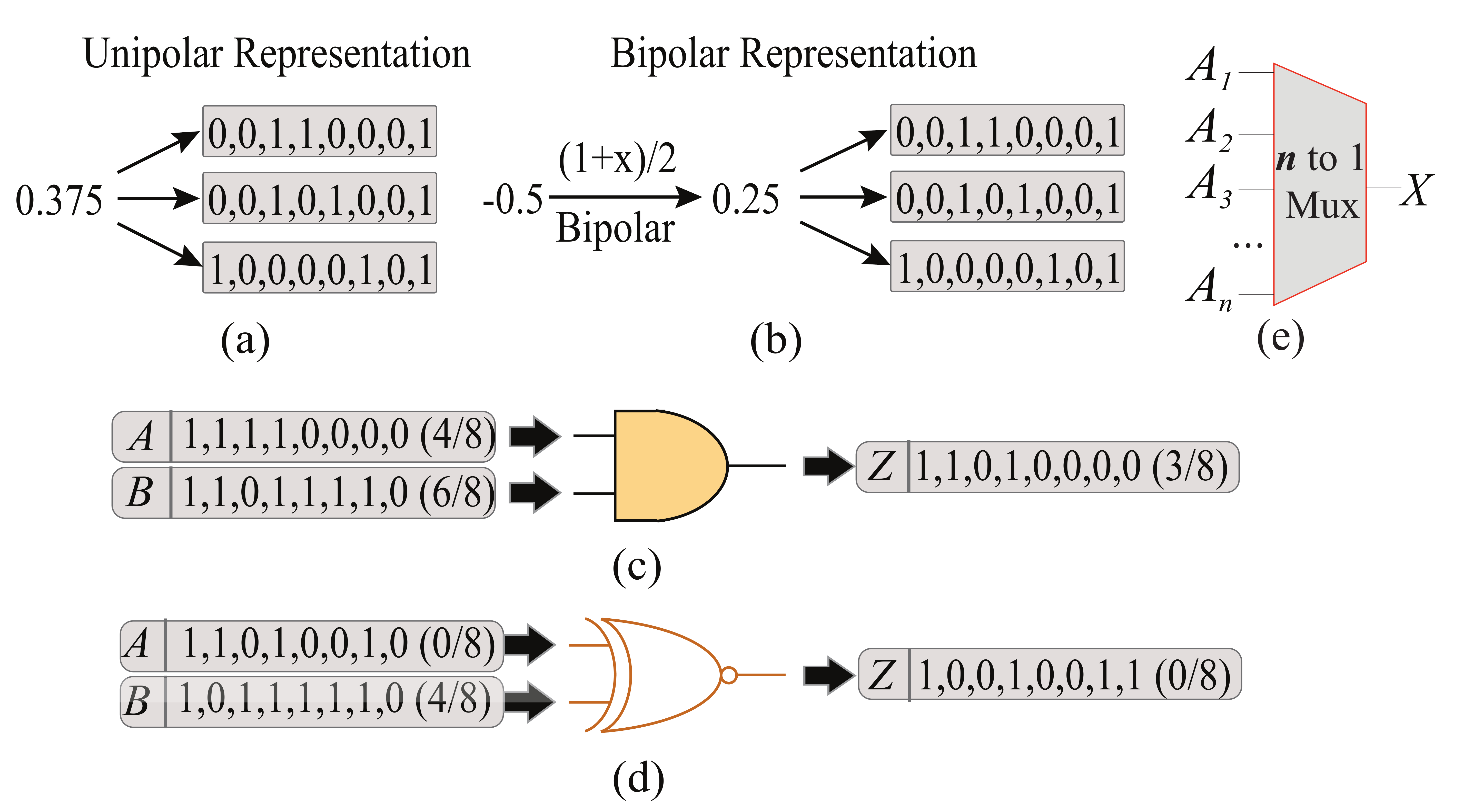}
	\caption{(a) Unipolar encoding and (b) bipolar encoding. (c) AND gate for unipolar multiplication. (d) XNOR gate for bipolar multiplication. (e) MUX gate for addition.}
\label{fig_SCgate}
\end{center}
\end{figure}

Stochastic computing (SC) is a paradigm that represents a number, named \emph{stochastic number}, by counting the number of ones in a bit-stream. For example, the bit-stream 0100110100 contains four ones in a ten-bit stream, thus it represents $x=P(X=1)=4/10=0.4$. In the bit-stream, each bit is independent and identically distributed (i.i.d.) which can be generated in hardware using stochastic number generators (SNGs). Obviously, the length of the bit-streams can significantly affect the calculation accuracy in SC \cite{ren2016designing}. 
In addition to this unipolar encoding format, SC can also represent numbers in the range of $[-1,1]$ using the bipolar encoding format. In this scenario, a real number $x$ is processed by $P(X=1)=(x+1)/2$. Thus 0.4 can be represented by 1011011101, as $P(X=1)=(0.4+1)/2=7/10$. -0.5 can be represented by 10010000, as it shown in Figure \ref{fig_SCgate}(b), with $P(X=1)=(-0.5+1)/2=2/8$.

Compared to conventional computing, the major advantage of stochastic computing is the significantly lower hardware cost for a large category of arithmetic calculations. 
A summary of the basic computing components in SC, such as multiplication and addition, is shown in Figure \ref{fig_SCgate}. As an illustrative example, a unipolar multiplication can be performed by a single AND gate since $P(A\cdot B=1)=P(A=1)P(B=1)$, and a bipolar multiplication is performed by a single XNOR gate since $c=2P(C=1)-1=2(P(A=1)P(B=1)+P(A=0)P(B=0))-1=(2P(A=1)-1)(2P(B=1)-1)=ab$. 

Besides multiplications and additions, SC-based activation functions are also developed \cite{ren2017sc}\cite{brown2001stochastic}. As a result, SC has become an interesting and promising approach to implement DNNs  \cite{sim2017new}\cite{ardakani2017vlsi}\cite{lee2017energy} with high performance/energy efficiency and minor accuracy degradation.

\section{Motivation and Challenges of the Proposed Work}
One observation we make is that SC technique, together with extremely small hardware footprint compared with binary computing, is especially compatible with the deep-pipelining nature in AQFP circuits. 
This is because SC uses a time-independent bit sequence for value representation, and there are no data dependency in a number of consecutive clock cycles (equal to the bit-stream length, ranging from 128 to 2,048). In this way the RAW hazards can be avoided without bubble insertions, which is not avoidable in conventional binary computing implemented using AQFP. 
 Additionally, it is important to note that the overhead of RNG, which is a significant overhead in SC using CMOS logic (even accounts for 40\%-60\% hardware footprint), can be largely mitigated in AQFP. This is because of extremely high efficiency in true RNG implementation (instead of pseudo-RNG) in AQFP, thanks to its unique operating nature.
 
\begin{figure}[t]
\centering
\includegraphics[width=0.4\textwidth]{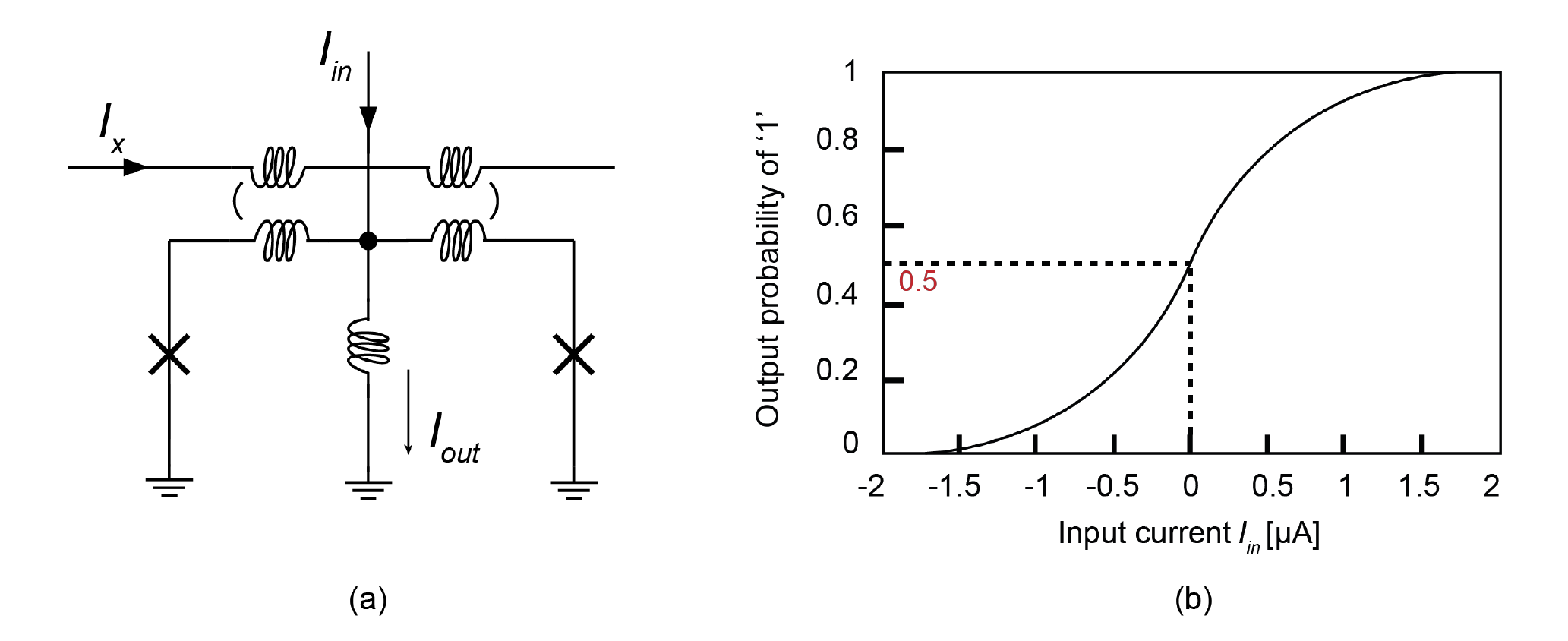}
\caption{(a). 1-bit true RNG in AQFP; (b). Output distribution (which will converge to 0 and 1).}
\label{fig:aqfp_rng}
\end{figure}

As illustrated in Fig. \ref{fig:aqfp_rng} (a), a 1-bit true RNG can be implemented using the equivalent circuit of an AQFP buffer. The direction of the output current $I_{out}$ represents the logic state 1 or 0. $I_{out}$ will be determined by the input current $I_{in}$. 
However, when $I_{in}=0$, the output current $I_{out}$ will be randomly 0 or 1 depending on the thermal noise, as shown in Fig. \ref{fig:aqfp_rng} (b). As a result, an on-chip RNG is achieved with two JJs in AQFP, and an independent random bit (0 or 1) will be generated in each clock cycle.

In a nutshell, we forecast SC to become a competitive technique for AQFP logic on a wide set of applications. For the acceleration of DNN inference phase in particular, SC has been demonstrated as a competitive technique to reduce hardware footprint while limiting accuracy loss \cite{alaghi2018promise}\cite{ren2017sc}. Compared with precise computing applications \cite{ardakani2017vlsi}, SC is more suitable for approximate computing where DNN is an example. Moreover, it has been proved recently on the equivalence between SC-based DNN and binary neural networks (BNN) \cite{wang2018universal}. As the latter originates from the deep learning society \cite{courbariaux2016binarized} and many accuracy enhancement techniques have been developed, these advances can be migrated to enhance the accuracy in SC-based DNNs to be close to the software-based, floating point accuracy levels.
\begin{figure}[t]
\begin{center}
\includegraphics[width = 0.32\textwidth]{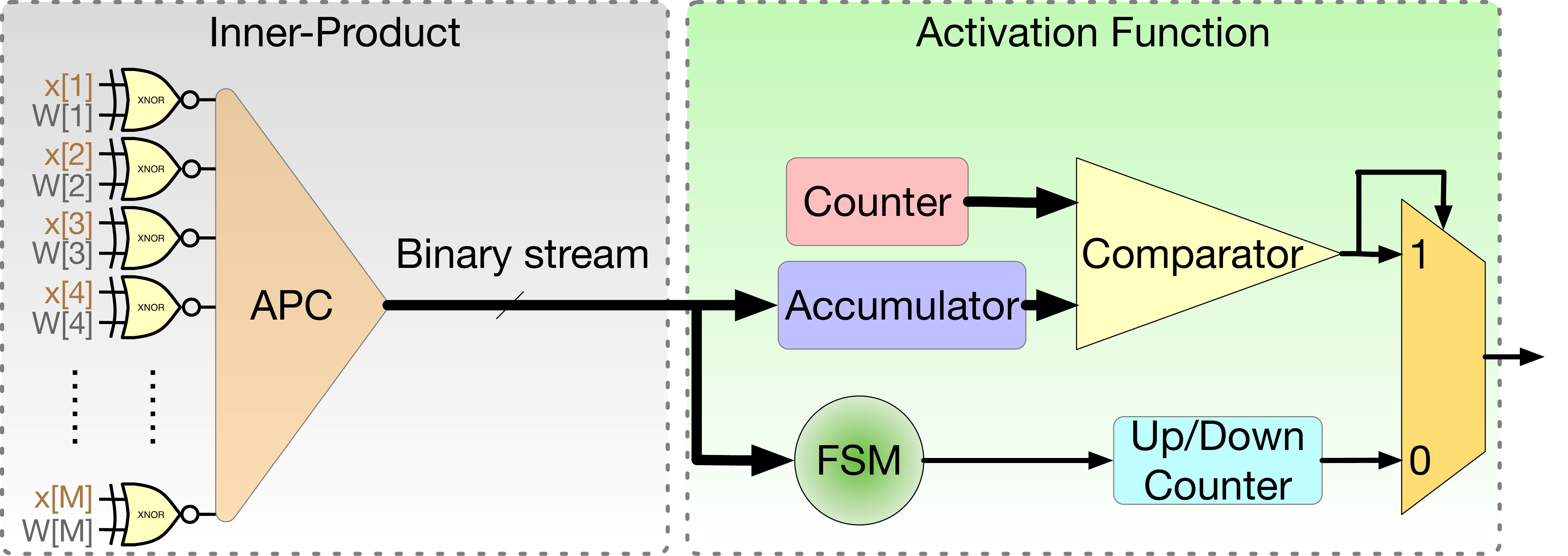}
	\caption{Feature extraction block of CMOS-based SC-based DNNs implementation in prior work.}
\label{fig:scdcnn}
\end{center}
\end{figure}

Despite the above advantages in SC-based DNNs using AQFP, certain challenges need to be overcome to realize an efficient implementation. 
One major challenge is the difficulty to implement accumulators/counters due to the super-deep pipelining nature.
This is because one single addition takes multiple clock phases, say $n$, in the deep-pipelining structure. Then the accumulation operation can only be executed once in every $n$ clock phases to avoid RAW hazards. Throughput degradation will be resulted in especially when a large number is allowed in the accumulator (in the case of baseline structure of SC-based DNNs\cite{ren2017sc} as we shall see later). Similarly, it is also challenging for efficient implementation of finite state machines (FSM) in AQFP.

\begin{figure}[b]
\begin{center}
\includegraphics[width = 0.45\textwidth]{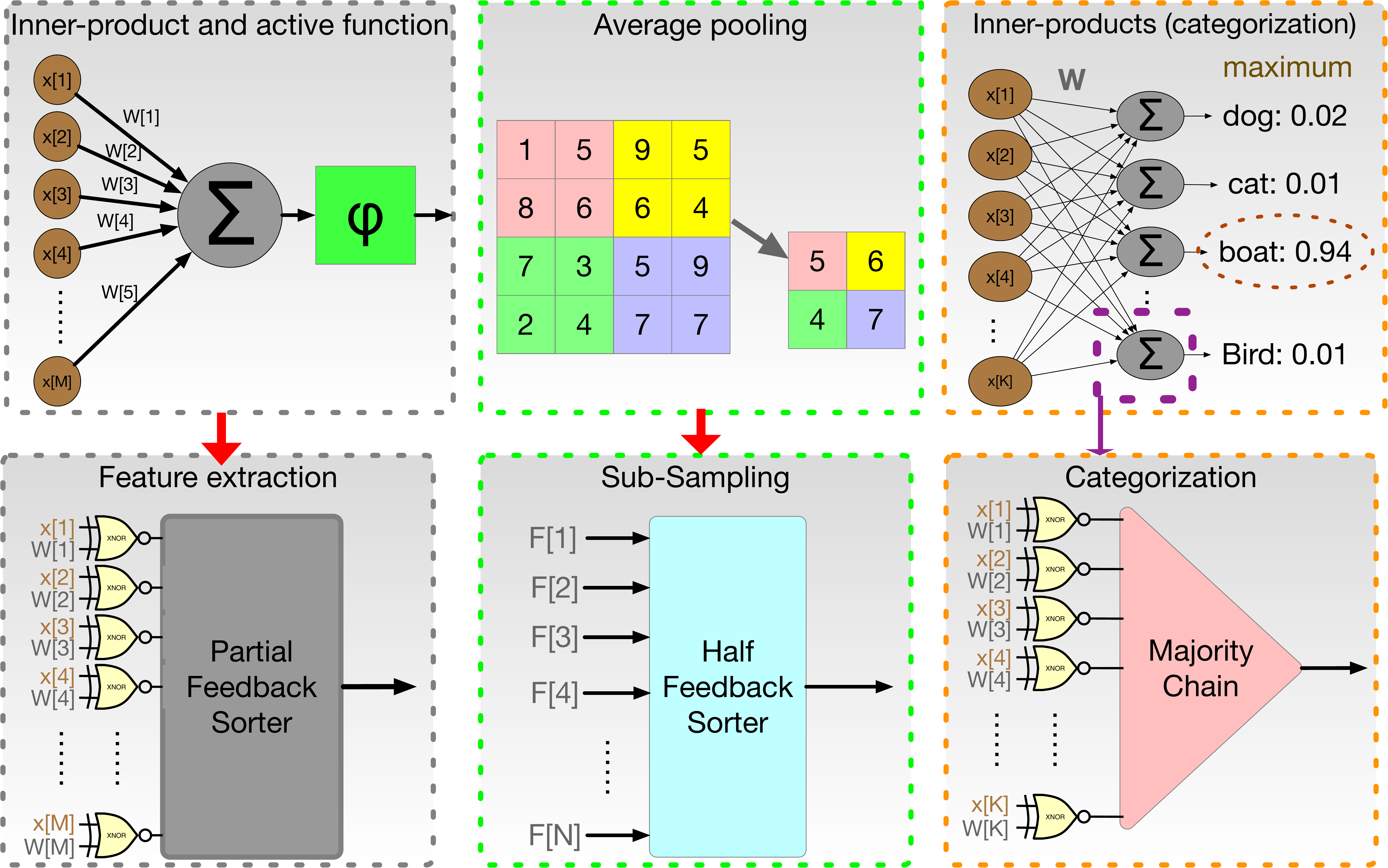}
	\caption{Proposed SC-based DNN architecture using AQFP.}
\label{fig:arch}
\end{center}
\end{figure}

\section{Proposed SC-based DNN Framework using AQFP}
\subsection{System Architecture: Difference from the Prior Arts}
Fig. \ref{fig:scdcnn} demonstrates an $M$-input \emph{feature extraction block} as defined in the prior work \cite{ren2017sc} of SC-based DNNs (built in CMOS), as a basic building block in CONV layers for feature extraction and FC layers for categorization. This structure performs inner product and activation calculation in the SC domain. The whole SC-based DNN chip comprises parallel and/or cascaded connection of such structures: the former for computing multiple outputs within a DNN layer, while the latter for cascaded DNN layers.

The functionality in the feature extraction block is to implement $y_i=\psi({\bf w}^T_i{\bf x}+b_i)$, where $\bf x$ and ${\bf w}_i$ are input vector and one weight vector, respectively; $b_i$ is a bias value; and $\psi$ is the activation function. In CMOS-based implementation of feature extraction block \cite{ren2017sc}, multiplication in the inner product calculation is implemented using XNOR gates (both weights and inputs are bipolar stochastic numbers); summation is implemented using \emph{approximate parallel counter} (APC) for higher accuracy or using mux-tree (adder-tree) for low hardware footprint; activation is implemented using binary counter (for APC output) or FSM (for mux-tree output).

The aforesaid challenge in accumulator/counter and FSM implementations in AQFP translates into the difficulty in the direct, AQFP-based implementation of the above feature extraction block. More specifically, the binary counter or FSM that are necessary for activation function implementation can only be executed once in multiple clock cycles to avoid RAW hazards, resulting in throughput degradation. This limitation needs to be overcome in our proposed SC-based DNN architecture using AQFP.

Fig. \ref{fig:arch} demonstrates our proposed SC-based DNN architecture using AQFP, which uses different implementation structures of inner product/activation for CONV layers and FC layers. For CONV layers, the number of inputs for inner product computation is less compared with that for FC layers. Meanwhile, the requirement of accumulation operation is more significant (in terms of effect on overall accuracy) in the CONV layers compared with FC layers. Based on these observations, we propose (i) accurate integration of summation and activation function in CONV layers using bitonic sorting network and feedback loop, and (ii) low-complexity categorization block for FC layers based on chain of majority gates. The former is more accurate and more complicated, which is compatible with CONV layers which have higher impact on overall accuracy but with smaller number of inputs. The latter has low complexity, and is compatible with FC layers which has lower impact on overall accuracy but with large number of inputs. The overall accuracy loss will be minor and under control as shall be shown in the experiments.


In the following subsections, we will describe in details in the proposed feature extraction block for CONV layers and categorization block for FC layers. 
Despite the distinct structure, both blocks essentially implement the inner product and activation $y_i=\psi({\bf w}^T_i{\bf x}+b_i)$ as discussed before, and multiplication is implemented in XNOR in SC domain. The inputs of both blocks are from previous DNN layer or from primary inputs, and are stochastic numbers in principle. 
The weights are stored (hardwired) in the AQFP chip in binary format for reducing hardware footprint, and will be transformed to stochastic format through RNG and comparators. In the following, we also describe (i) the stochastic number generator that can be implemented with ultra-high efficiency in AQFP, and (ii) accurate sub-sampling block in AQFP for pooling operation, which has higher accuracy than the version in \cite{ren2017sc}.

\begin{figure}[b]
\centering
\includegraphics[width=0.2\textwidth]{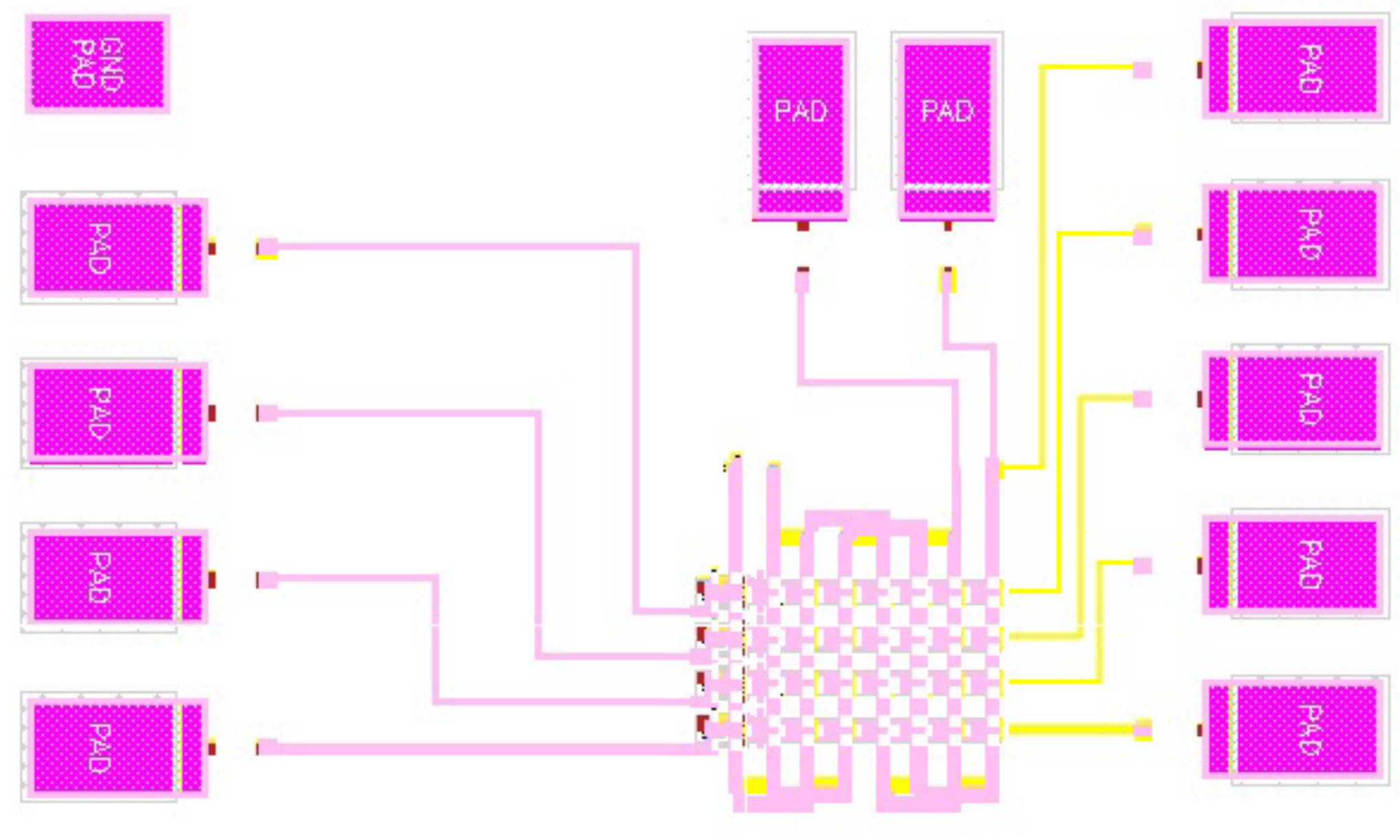}
\caption{Layout of 1-bit true RNG using AQFP.}
\label{fig:rng_layout}
\end{figure}

A \emph{stochastic number generator} (SNG) is a key part in the SC scheme, which converts a binary number to its corresponding stochastic format. This is achieved via comparing between the binary number (to convert) with a stream of random numbers \cite{gaines1969stochastic}\cite{alaghi2013survey}, and a uniformly distributed RNG is needed for this conversion process. In AQFP, an $n$-bit true RNG can be implemented using $n$ 1-bit true RNGs, whose layout is shown in Fig. \ref{fig:rng_layout}, each using two JJs for efficient implementation as discussed in the previous section.   
An SNG can be implemented using the $n$-bit true RNG and $n$-bit comparator, where $n$ is the number of bits of the source binary number. 

\begin{figure}[t]
\centering
\includegraphics[width=0.3\textwidth]{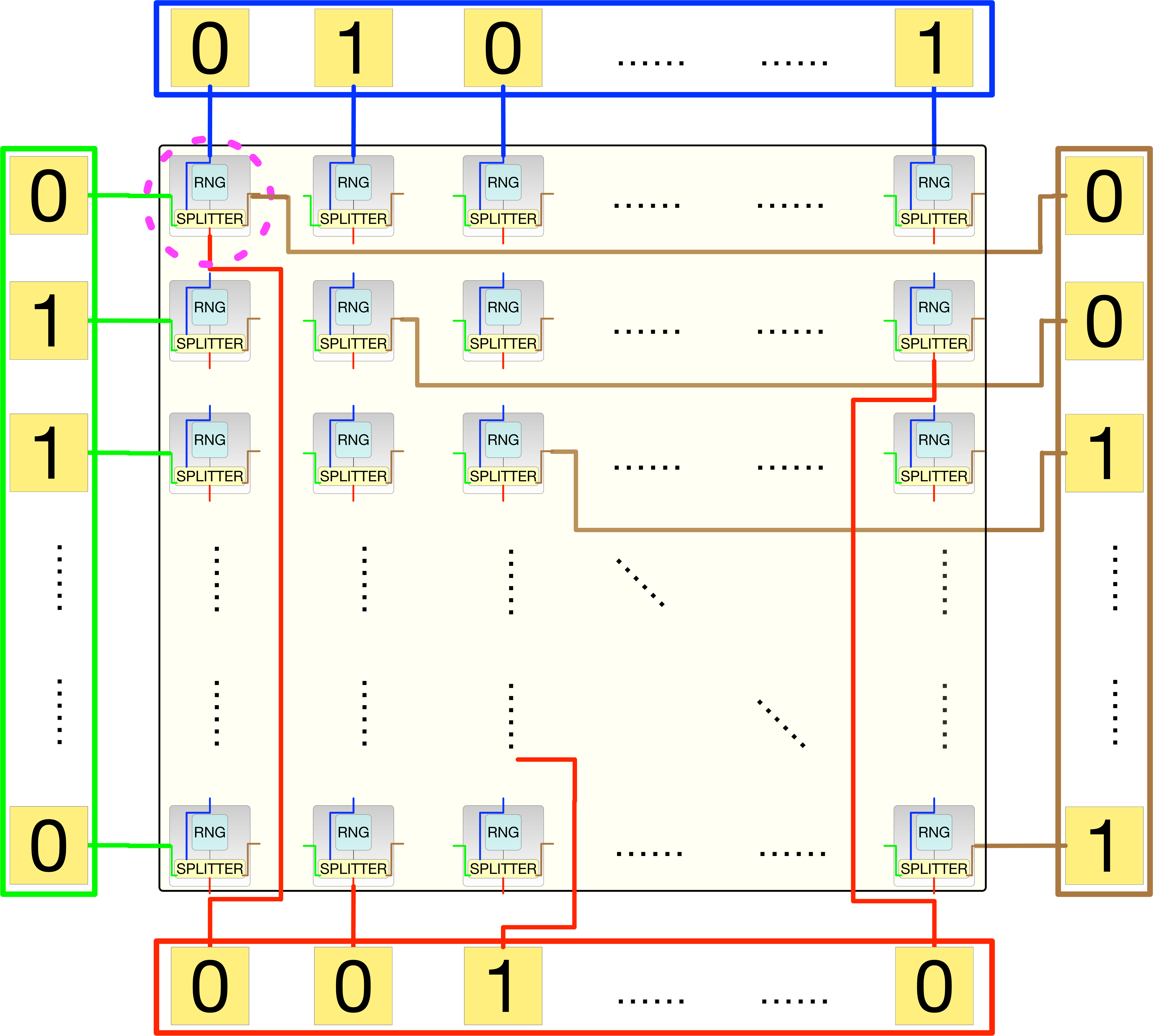}
\caption{True RNG cluster consisting of $N \times N$ unit true RNGs, where each unit is shared by four $N$-bit random numbers.}
\label{fig:rng_cluster}
\end{figure}

For more efficient hardware utilization, we propose a true random number generator matrix, in which each true RNG unit can be exploited in more than one SNGs. As shown in Fig. \ref{fig:rng_cluster}, an $N \times N$ RNG matrix is capable of generating $4N$ $N$-bit random numbers in parallel. Each true RNG unit is shared among four random numbers. This design can maintain limited correlation because all unit RNGs are independent and each two output random numbers only share a single bit in common.  

\subsection{Integration of Summation and Activation Function in CONV Layers}
To overcome the difficulty in accumulator implementation in AQFP, we re-formulate the operation of SC-based feature extraction block in a different aspect.
The stochastic number output of inner-product (and activation), $\mathbf{SO}$, should reflect the summation of input-weight products, whose stochastic format can be viewed as a matrix $\mathbf{SP}$. Each row of $\mathbf{SP}$ is the stochastic number/stream of a input-weight product. Therefore, the binary value represented by $\mathbf{SO}$ is:
\begin{multline}
     \dfrac{2\times{\sum_{i=1}^{N}\mathbf{SO}_i} - N}{N} \\= clip\Big(\dfrac{2\times\sum_{i=1}^{N}\sum_{j=1}^{M}\mathbf{SP}_{i,j} - N \times M}{N}, -1, 1\Big)
\end{multline}
where $N$ is the length of the stochastic stream and $M$ is the number of inputs. The $clip$ operation restricts the value between the given bounds. This formulation accounts for inner product (summation) and activation function. Consequently, 
\begin{equation}
    \sum_{i=1}^{N}\mathbf{SO}_i = clip\Big(\sum_{i=1}^{N}\sum_{j=1}^{M}\mathbf{SP}_{i,j} - \dfrac{M-1}{2}\times N, 0, N\Big)
\end{equation}
That is, the total number of 1's in the output stochastic stream, $\mathbf{SO}$, should equal to the total number of 1's in the input matrix, $\mathbf{SP}$, minus $\dfrac{M-1}{2}\times N$. The $n$-th bit of $\mathbf{SO}$ can be determined by the following value:
\begin{multline}
    \sum_{i=1}^{n}\sum_{j=1}^{M}\mathbf{SP}_{i,j} - i\times\dfrac{M-1}{2}-\sum_{i=1}^{n-1}\mathbf{SO}_{i}
    \\=\sum_{i=1}^{n-1} {\big(\sum_{j=1}^{M}\mathbf{SP}_{i,j} - \dfrac{M-1}{2}-\mathbf{SO}_{i}\big)} + \sum_{j=1}^{M}\mathbf{SP}_{n,j} - \dfrac{M-1}{2}
\end{multline}
More precisely, the $n$-th bit of $\mathbf{SO}$ is 1 if the above value is greater than 0, and is 0 otherwise. The above calculation needs to accumulate $\sum_{j=1}^{M}\mathbf{SP}_{i,j} - {(M-1)}/{2}-\mathbf{SO}_{i}$, further denoted as $\mathbf{D}_i$, in each clock cycle. To simplify the computation given hardware limitations, we propose an effective, accurate approximation technique as shown in Algorithm 1.

\begin{algorithm}[b]\scriptsize			
	\SetKwInOut{Input}{input}\SetKwInOut{Output}{output}
	
	\Input{$\mathbf{SP}$ is the matrix containing all input-weight products and bias \\
	       $N$ is the bit-stream length \\
	       $M$ is the input size\\
	       }
	\Output{$\mathbf{SO}$ is the activated inner-product.}
		
	$\mathbf{D_{prev}} = \mathbf{0}$;  \ $//$initialize feed back vector to all 0 \\		
	\For{$i++ < N$}{		
		$\mathbf{D_i} = \mathbf{SP}[:i]$; \ $//$current column \\
		$\mathbf{D_s} = sort((D_i, D_{prev}), descending)$; \ $//$sort the vector consist of the current column and previous feedback in descending order \\ 
		$\mathbf{SO}[i] = \mathbf{D_s}[(M-1)/2]$; \ $//$  check the $(M-1)/2$-th bit for output as the top $(M-1)/2$ bits are subtracted each iteration \\
    	$\mathbf{D_{prev}} = \mathbf{D_i}[(M+1)/2: (M+1)/2+M]$; \ $//$ feedback the $M$ bits following the $(M-1)/2$-th bit for next iteration \\
	}
\caption{\footnotesize Proposed sorter-based feature extraction block (implemented by SC in hardware)}\label{algo_fe}
\end{algorithm}

 As described in Algorithm \ref{algo_fe}, each column of the input bit-stream matrix $\mathbf{SP}$ (corresponding to the current clock cycle), combined with the remaining $M$ bits from the previous iteration, are fed to the sorting block. By performing binary sorting, the sorting block will decide whether the number of 1's in all inputs is greater than $(M-1)/2$ or not. More specifically, the $(M-1)/2$-th bit of the sorted results can be used to determine the next stochastic bit of $\mathbf{SO}$ (to be 1 or 0). 
 The $M$ bits following the $(M-1)/2+1$-th bit at output, whose number of 1's represents $\sum_{i=1}^{n} clip({\bf{D}}_i, 0, 1)$, are fed back to the sorting block. The above step acts as subtraction as the top $(M-1)/2$ bits are ignored, and the surplus 1's can still be used in the following iteration. 

\begin{figure}[t]
\centering
\includegraphics[width=0.38\textwidth]{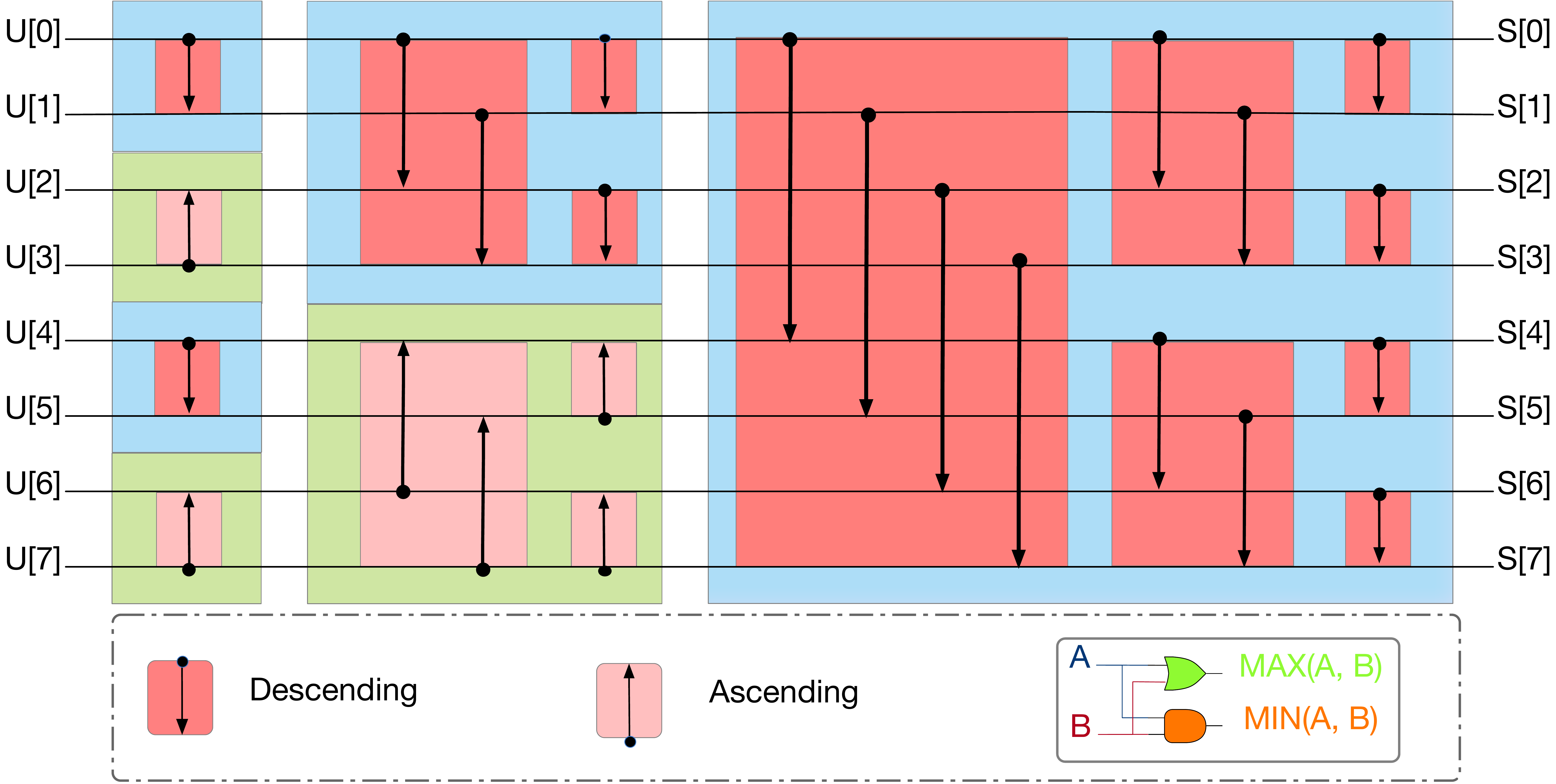}
\caption{Example of 8-input binary bitonic sorter.}
\label{fig:bitonic_sorter_example}
\end{figure}

\begin{figure}[t]
\centering
\includegraphics[width=0.38\textwidth]{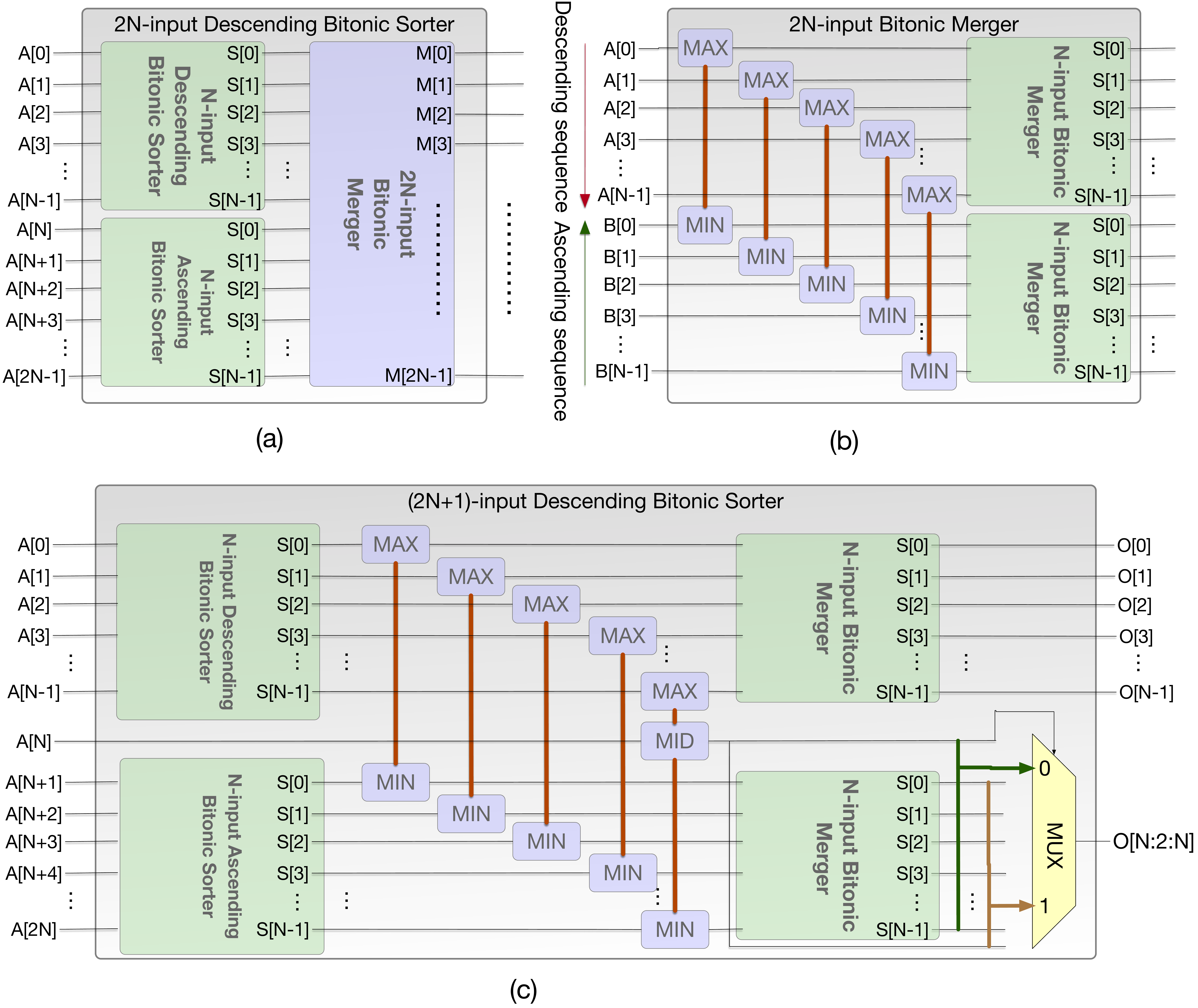}
\caption{(a) Bitonic sorter for even-numbered inputs. (b) Bitonic merger for even-numbered inputs. (c) Bitonic sorter for odd-numbered inputs.}
\label{fig:bitonic_sorter}
\end{figure}

The binary sorting operation can be realized using the bitonic sorting network \cite{liszka1993generalized}, which applies a divide-and-merge technique. Fig. \ref{fig:bitonic_sorter_example} shows an 8-input binary bitonic sorter, where each sorting unit can be implemented using an AND gate for the maximum and an OR gate for the minimum of the two input bits. A bitonic sorter has two stages: sorting the top and bottom halves of its inputs where the bottom half is sorted in the opposite order of the top half; then merge the two sorted parts. As shown in Fig. \ref{fig:bitonic_sorter} (a), a binary bitonic sorter with even inputs can be built up with bitonic sorters with 1/2 size and a bitonic merger, which is shown in Fig. \ref{fig:bitonic_sorter} (b). For a binary bitonic sorter with odd inputs, we propose to add a three-input sorter in the first merging stage to sort the maximum of the bottom half, the middle input, and the minimum of the top half. The maximum of the three is fed to the top merger; the minimum is feed to the bottom merger; and the median, which is also the median of the entire inputs or the minimum, is used as a select signal to decide the bottom half output. This design allows any binary bitonic sorter to be implemented in a modular manner. The three-input sorter can be implemented using an AND gate, an OR gate and an majority gate for the median. 

\begin{figure}[t]
\centering
\includegraphics[width=0.37\textwidth]{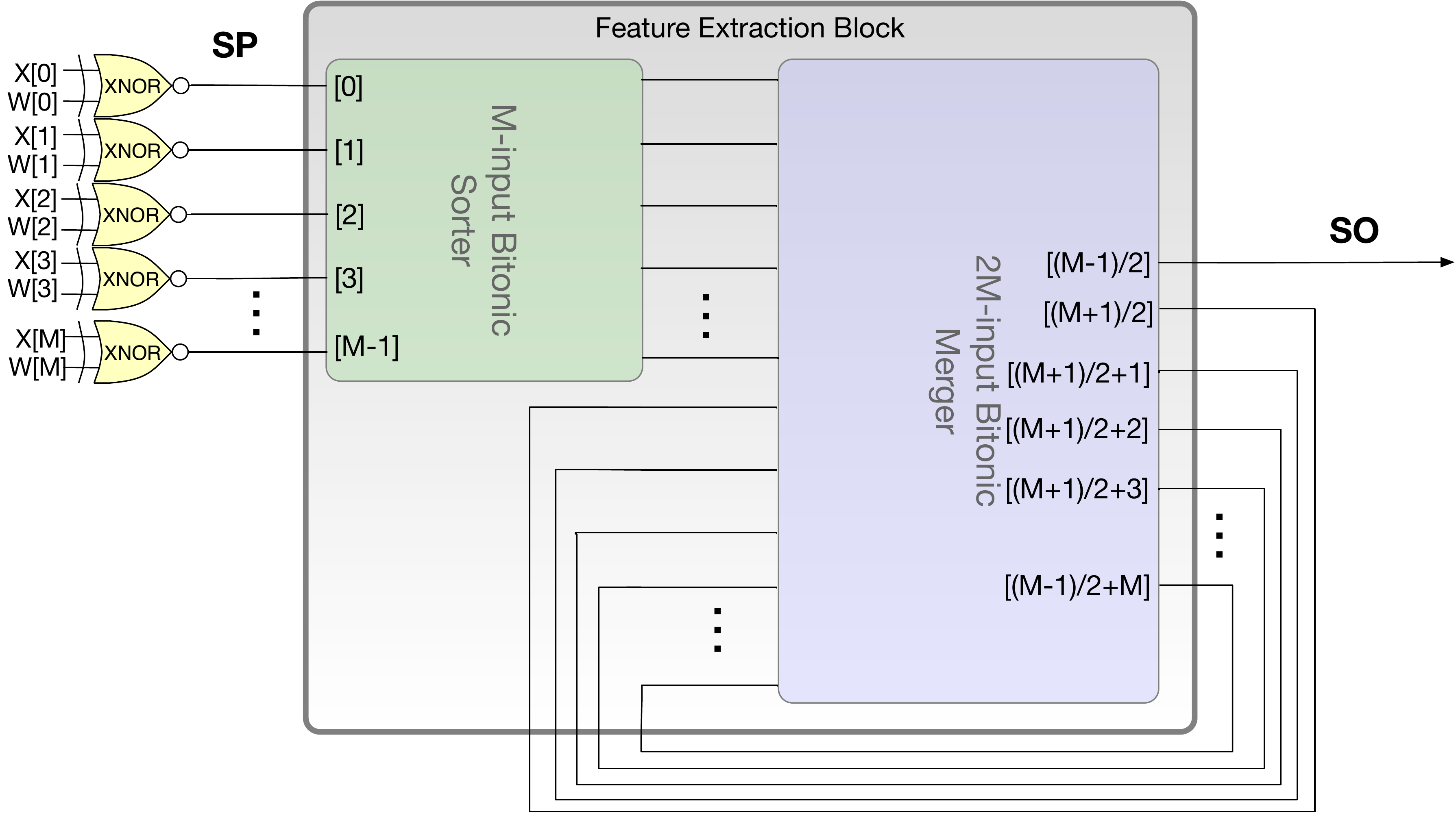}
\caption{Proposed sorter based feature-extraction block.}
\label{fig:feat_ex}
\end{figure}

As shown in Fig. \ref{fig:feat_ex}, the operation in (3) can be realized using a bitonic sorting circuit. As the feedback vector is already sorted, only the input column needs to be sorted. The two sorted sequences are then merged using a bitonic merger to form the final sorted vector. When $M$ is an even number, a neutral noise of repeated 0 and 1 stochastic sequence (whose value is 0) is added to the input, in order to mitigate the limitation that $(M-1)/2$ cannot be fully represented in an integer format in this case.  
 
\begin{figure}[b]
\centering
\includegraphics[width=0.2 \textwidth]{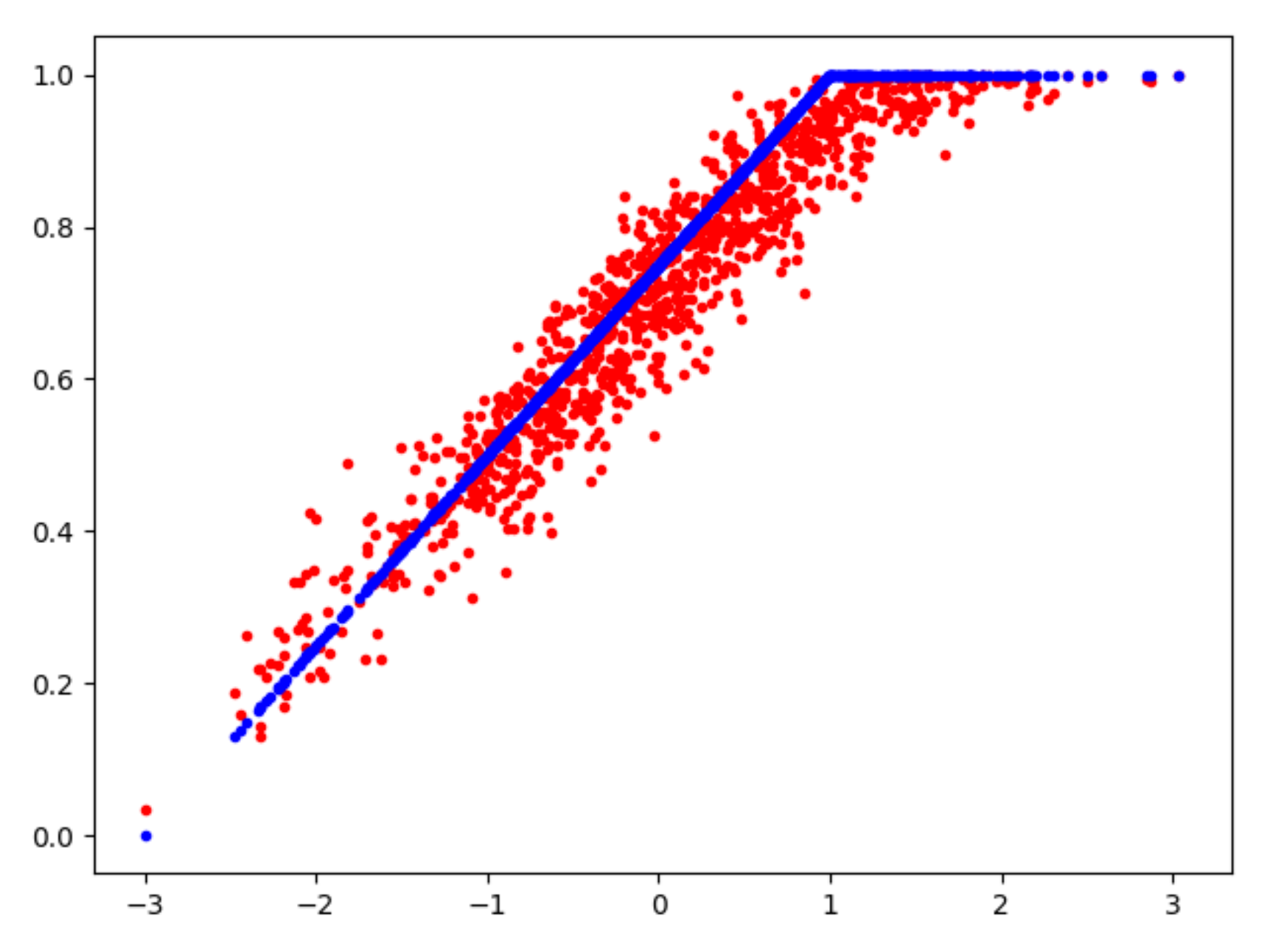}
\caption{Activated output of the proposed feature extraction block.}
\label{fig:feature_extract_example}
\end{figure}

\begin{table}[b]
\centering
\caption{Absolute inaccuracy of the bitonic sorter-based feature extraction block.}
\label{tbl:acc_fe}
{
\begin{tabular}{|c|c|c|c|c|c|}
\hline
\multirow{1}{*}{Input size} &\multicolumn{5}{c|}{Bit-stream length}\\ 
\cline{2-6}
&128 &256 &512 &1024 &2048\\
\hline
9 &0.1131 &0.0847 &0.0676 &0.0573 &0.0511\\
\hline
25 &0.1278 &0.0896 &0.0674 &0.0536 &0.0434\\
\hline
49 &0.1267 &0.0954 &0.0705 &0.0528 &0.0468\\
\hline
81 &0.129 &0.0937 &0.0685 &0.0531 &0.0396\\
\hline
121 &0.1359 &0.0942 &0.0654 &0.0513 &0.0374\\
\hline
\end{tabular}
}
\end{table}

As shown in Fig. \ref{fig:feature_extract_example}, the output of the proposed feature extraction block resembles a shifted rectified linear unit (ReLU). Table \ref{tbl:acc_fe} shows the absolute error of the proposed bitonic sorter based feature extraction block. The performance of the proposed sorter-based feature extraction block is very consistent, and does not degrade as the input size increases. The inaccuracy is no more than 0.06 for bit-streams longer than 1024.

\subsection{Sub-Sampling Block in AQFP}
The sub-sampling operation is realized using the pooling block. The max-pooling operation is not compatible with AQFP as it requires FSM for implementation, while the multiplexer based average pooling implementation \cite{ren2017sc} suffers from high inaccuracy as input size grows. 

We address these limitations and propose an accurate, AQFP-based average pooling block. The output stochastic number/stream of the average-pooling block, $\mathbf{SO}$, should satisfy:
\begin{multline}
     \dfrac{2\times{\sum_{i=1}^{N}\mathbf{SO}_i} - N}{N} \\= clip\Big(\dfrac{2\times\sum_{i=1}^{N}\sum_{j=1}^{M}\mathbf{SP}_{i,j} - N \times M}{N\times M}, -1, 1\Big)
\end{multline}
Therefore:
\begin{equation}
    \sum_{i=1}^{N}\mathbf{SO}_i = clip\Big(\dfrac{\sum_{i=1}^{N}\sum_{j=1}^{M}\mathbf{SP}_{i,j}}{M}, 0, N\Big)
\end{equation}

\begin{algorithm}[t]\scriptsize			
	\SetKwInOut{Input}{input}\SetKwInOut{Output}{output}
	
	\Input{$\mathbf{SP}$ is the matrix containing all inputs \\
	       $N$ is the bit-stream length \\
	       $M$ is the input size\\
	       }
	\Output{$\mathbf{SO}$ is the average of its inputs.}
		
	$\mathbf{D_{prev}} = \mathbf{0}$;  \ $//$initialize feed back vector to all 0 \\		
	\For{$i++ < N$}{		
		$\mathbf{D_i} = \mathbf{SP}[:i]$; \ $//$current column \\
		$\mathbf{D_s} = sort((D_i, D_{prev}), descending)$; \ $//$sort the vector consist of the current column and previous feedback in descending order \\ 
		\eIf{$\mathbf{D_s}[M] == 1$}
	        {$\mathbf{D_{prev}} = \mathbf{D_s}[1:M]$;\ $//$ less than $M$ 1s have been encountered since last 1 has been output, feedback current saved 1s for next iteration}
	        {$\mathbf{D_{prev}} = \mathbf{D_s}[M+1:2M]$;\ $//$ more than $M$ 1s encountered, output should be 1 and feedback the following surplus $M$ bits}
		$\mathbf{SO}[i] = \mathbf{D_s}[M]$; \ $//$   \\
	}
\caption{\footnotesize Proposed sorter-based average-pooling block (implemented by SC in hardware)}\label{algo_ap}

\end{algorithm}
\begin{figure}[t]
\centering
\includegraphics[width=0.37\textwidth]{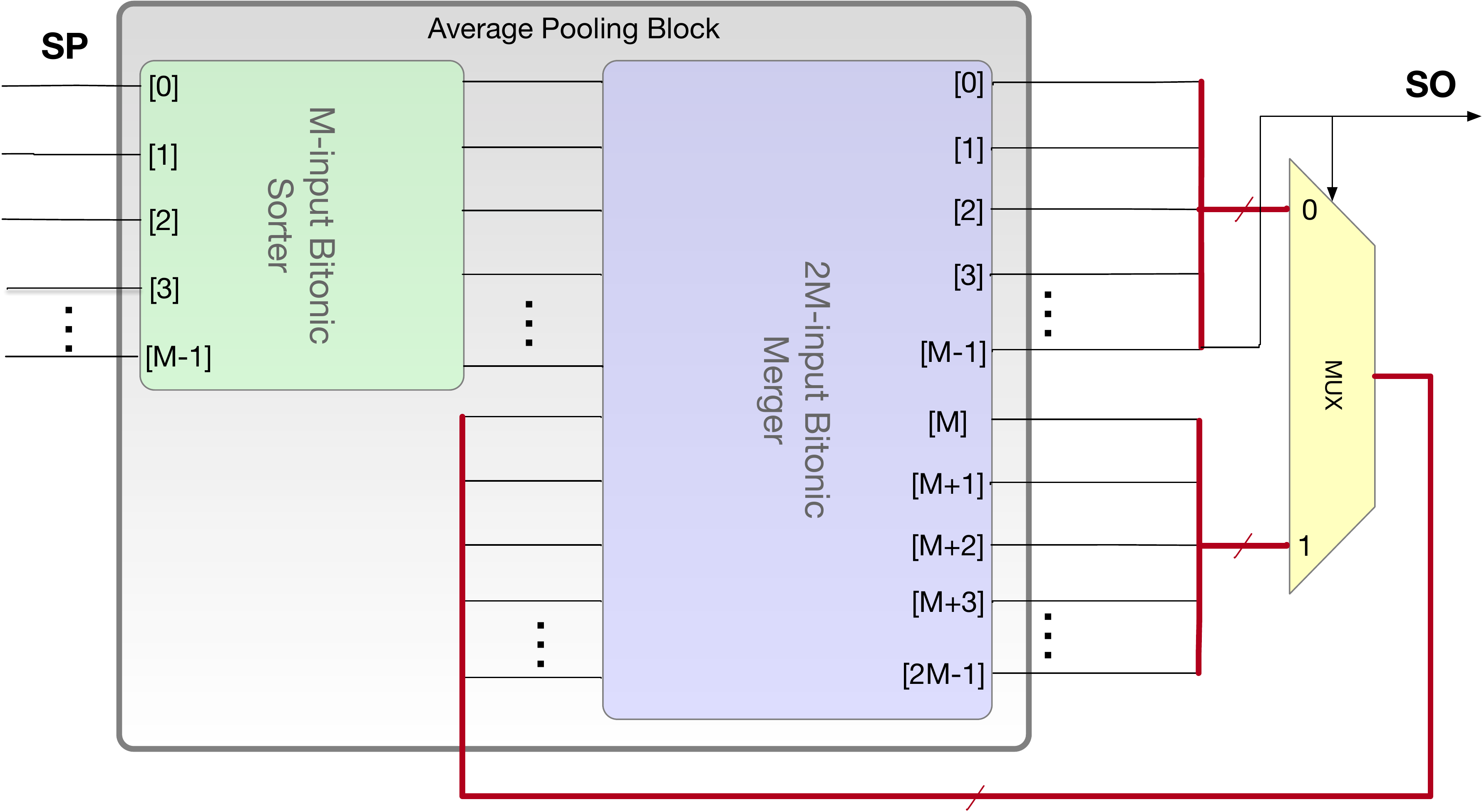}
\caption{Proposed bitonic sorter based sub-sampling block.}
\label{fig:sub_samp}
\end{figure}

In other words, the number of 1's in $\mathbf{SO}$ should be equal to the number of 1's in $\mathbf{SP}$ divided by $M$ (the number of inputs). 
Using the similar idea as the proposed feature extraction block, the sub-sampling block can also be realized using a sorting circuit. 

\begin{table}[b]
\centering
\caption{Absolute inaccuracy of the bitonic sorter-based average-pooling block.}
\label{tbl:acc_ap}
{
\begin{tabular}{|c|c|c|c|c|c|}
\hline
\multirow{1}{*}{Input size} &\multicolumn{5}{c|}{Bit-stream length}\\ 
\cline{2-6}
&128 &256 &512 &1024 &2048\\
\hline
4 &0.0249 &0.0163 &0.0115 &0.0085 &0.0058\\
\hline
9 &0.0173 &0.0112 &0.0079 &0.0055 &0.0039\\
\hline
16 &0.0141 &0.0089 &0.0061 &0.0042 &0.0030\\
\hline
25 &0.0122 &0.0078 &0.0049 &0.0033 &0.0024\\
\hline
36 &0.0105 &0.0065 &0.0043 &0.0029 &0.0019\\
\hline
\end{tabular}
}
\end{table}

As described in Algorithm \ref{algo_ap}, the current column of inputs and previous feedback are sorted to group all 1's since the last 1 produced in $\mathbf{SO}$. The $M$-th bit is used to determine the next output bit and feedback bits. If a 1 is produced, the top $M$ 1's are discarded and the remaining are feedback to the input. Otherwise, the top $M$ bits are feedback for further accumulation. In summary, this would allow one 1 to be produced in $\mathbf{SO}$ for every $M$ 1's in $\mathbf{SP}$. 

The average pooling block is also implemented using a bitonic sorter, as shown in Fig. \ref{fig:sub_samp}. Similar to the design of the feature extraction block, only one $M$-input sorter is needed to sort the input column as the feedback vector is already sorted. The final sorted vector can be generated using a bitonic merger to merge the sorted input column and previous feedback. The output bit also acts as the select signal for feedback vector. 

The proposed sub-sampling block produces very low inaccuracy as presented in Table \ref{tbl:acc_ap}, which is usually far less than 0.01 for bit-streams longer than 1024. Even when the input size is very small, the proposed bitonic sorter-based average pooling block can still operate with negligible error, which is no more than 0.025 regardless of the bit-stream length.   
\vspace{-1em}
\subsection{Categorization Block for FC Layers in AQFP}
Categorization blocks are for (final) FC layers of DNNs, in which each output is the inner-product of corresponding input and weight vectors. 
We find it not ideal to use the same implementation for the categorization block as feature-extraction block for CONV layers, as the categorization block usually involves more inputs, leading to higher hardware footprint.
In addition, the categorization result is reflected via the ranking of the outputs rather than relying on the calculation accuracy of inner product.
As a result of the above two reasons, we propose to implement the categorization block in a less complicated manner, which is capable of maintaining the relative ranking of outputs without offering precise inner-product.
\begin{figure}[t]
\centering
\includegraphics[width=0.32\textwidth]{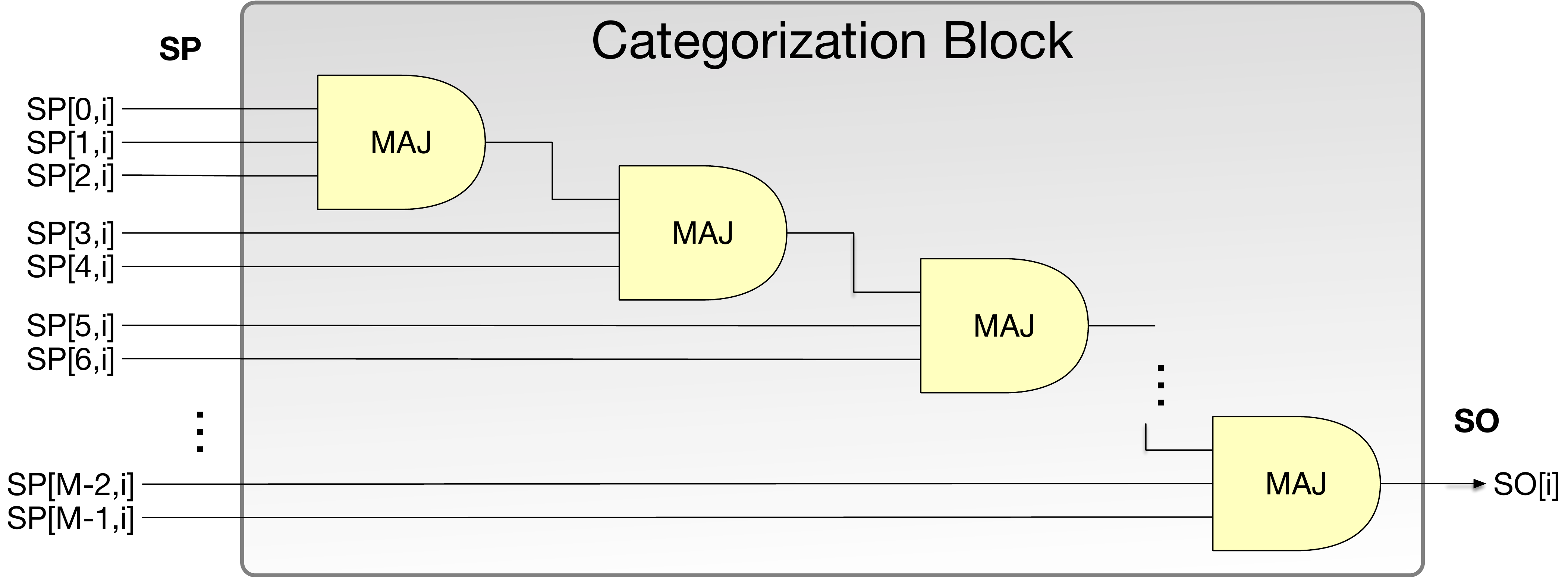}
\caption{Categorization block implementation using majority chain structure.}
\label{fig:maj_chain}
\end{figure}
\begin{table}[b]
\centering
\caption{Relative inaccuracy of the majority chain-based categorization block}
\label{tbl:acc_tp1}
\resizebox{\columnwidth}{!}{
\begin{tabular}{|c|c|c|c|c|c|}
\hline
\multirow{1}{*}{Input size} &\multicolumn{5}{c|}{Bit-stream length}\\ 
\cline{2-6}
&128 &256 &512 &1024 &2048\\
\hline
100 &0.3718\% &0.2198\% &0.1235\% &0.0620\% &0.0376\%\\
\hline
200 &0.2708\% &0.2106\% &0.1671\% &0.0743\% &0.0301\%\\
\hline
500 &0.2769\% &0.2374\% &0.1201\% &0.0687\% &0.0393\%\\
\hline
800 &0.2780\% &0.1641\% &0.1269\% &0.0585\% &0.0339\%\\
\hline
\end{tabular}
}
\end{table}

The proposed categorization block is implemented using the majority logic, whose output is the majority of all its inputs, i.e. the output is 1 if the input vector has more 1's than 0's; the output is 0 otherwise. The relative value/importance of each output can be reflected in this way, thereby fulfilling the requirement of categorization block in FC layers.The proposed categorization logic can be realized using a simple majority chain structure. Thanks to the nature of AQFP technology, a three-input majority gate costs the same hardware resource as a two-input AND/OR gate. As shown in Fig. \ref{fig:maj_chain}, a multi-input majority function can be factorized into multi-level of three-input majority gate chain based on the following equation: $Maj(x_0, x_1, x_2, x_4, x_5) = Maj(Maj(x_0, x_1, x_2), x_4, x_5)$.

Table. \ref{tbl:acc_tp1} shows the top 1 inaccuracy of the proposed categorization block. We simulate 10 categorization outputs with 100, 200, 500 and 800 different inputs under different lengths of bit-streams. The inaccuracy is evaluated using the relative difference between the highest output value in software and in SC domain. The relative inaccuracy can be limited to 0.4\%, i.e., if the largest output outscores the second largest by more than 0.4\%, the majority chain based categorization block can give the correct classification. As in most categorization cases, the highest output is usually far greater than the rest, the proposed majority chain categorization block should be able to operate with high accuracy. 
\begin{figure}[t]
\centering
\includegraphics[width=0.4\textwidth]{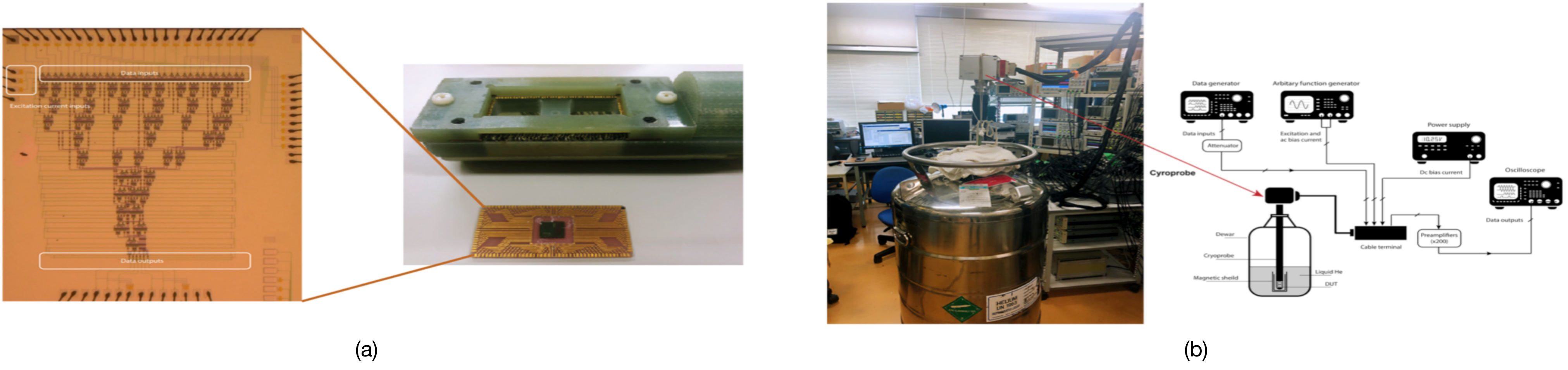}
\caption{AQFP chip testing.}
\label{fig:aqfp_testing}
\end{figure}

\section{Software Results and Hardware Prototype Testing}
The proposed SC-based DNN using AQFP is evaluated in various aspects: 1) component-wise hardware performance; 2) application-level performance; 3) system-level hardware utilization efficiency. For hardware performance of the four basic blocks, we compare their energy efficiency and latency with CMOS-based counterparts. The application-level performance is evaluated by testing the accuracy of the proposed framework on the MNIST \cite{lecun-mnisthandwrittendigit-2010} dataset. Finally, the overall hardware resource utilization for two different DNNs (one for the MNIST dataset and one with deeper architecture for more complex tasks) are compared with SC-based DNN implementations using CMOS. 

\begin{table}[t]
\centering
\caption{Hardware utilization of stochastic number generator}
\label{tbl:comp_sng}
\resizebox{\columnwidth}{!}{
\begin{tabular}{|c|c|c|c|c|}
\hline
\multirow{1}{*}{Output size}&\multicolumn{2}{c|}{Energy($pJ$)} &\multicolumn{2}{c|}{Delay($ns$)}\\ 
\cline{2-5}
&AQFP &CMOS &AQFP &CMOS\\
\hline
100 &9.700E-5 &14.42 &0.2 &0.6  \\
\hline
500 &4.850E-4 &72.11 &0.2 &0.6   \\
\hline
800 &7.760E-4 &115.4 &0.2 &0.6   \\
\hline
\end{tabular}
}
\end{table}

In addition we have verified the functionality of a feature extraction chip, as shown in Fig. \ref{fig:aqfp_testing} (a), which is fabricated with AIST $10kA/cm^2$ Niobium high-speed standard process (HSTP) \cite{takeuchi2017adiabatic}. The functionality of the chip is verified through 4.2K low-temperature measurements. As shown in Fig. \ref{fig:aqfp_testing} (b), the under-testing chip is embedded in a cryoprobe and inserted into a liquid Helium Dewar to cool down to 4.2K, with the protection of a double-layer magnetic shield. On-chip I/Os are connected to a cable terminal through the cryoprobe. A data pattern generator, an arbitrary function generator and a power supply are employed to generate the date inputs and excitation current. Oscilloscopes with pre-amplifiers are used to read the outputs.     
\begin{table}[b]
\centering
\caption{Hardware utilization of feature extraction block}
\label{tbl:comp_feb}
\resizebox{\columnwidth}{!}{
\begin{tabular}{|c|c|c|c|c|}
\hline
\multirow{1}{*}{Input size}&\multicolumn{2}{c|}{Energy($pJ$)} &\multicolumn{2}{c|}{Delay($ns$)}\\ 
\cline{2-5}
&AQFP &CMOS &AQFP &CMOS\\
\hline
9 &2.972E-4 &320.819 &2.2 &1024.0\\
\hline
25 &1.350E-3 &520.704 &3.4 &1228.8\\
\hline
49 &3.978E-3 &843.469 &4.8 &1535.0\\
\hline
81 &9.168E-3 &1099.776 &6.6 &1741.8\\
\hline
121 &1.333E-2 &2948.496 &6.8 &1946.6\\
\hline
500 &9.147E-2 &6807.552 &10.8 &2455.6\\
\hline
800 &0.186 &9804.800 &12.4 &2868.2\\
\hline
\end{tabular}
}
\end{table}

\subsection{Hardware Utilization}
To validate the efficiency of the proposed SC-based DNN using AQFP, we firstly compare components-wise and overall hardware utilization with CMOS-based implementation. The CMOS implementation is synthesized with 40nm SMIC CMOS process using Synopsys Design Compiler. We test and compare SNGs, feature extraction blocks, sub-sampling blocks and categorization blocks with physical different configuration.  

Table \ref{tbl:comp_sng} shows hardware utilization comparison for SNGs to generate 1024 bit-stream stochastic numbers based on 10-bit random numbers generated by RNGs. The AQFP based design is much efficient given its advantage in true random number generation in addition to the clusterized RNG design. The proposed AQPF SNG operates at $1.48\times 10^{5}$ times higher energy efficiency compared to its CMOS counterpart.    
The bitonic sorter based feature extraction block using AQFP is very energy efficient as its energy efficiency is $1.08\times10^6$ times higher than CMOS based implementation while being much faster thanks to its effective bitonic sorter-based architecture. For very large and dense layers, we still consider them as feature extraction layers as they derives global features. The sorter-based implementation, regardless of the big input size, can still maintain reasonable hardware performance. 

\begin{table}[t]
\centering
\caption{Hardware utilization of sub sampling block}
\label{tbl:comp_sub}
\resizebox{\columnwidth}{!}{
\begin{tabular}{|c|c|c|c|c|c|c|}
\hline
\multirow{1}{*}{Input size}&\multicolumn{2}{c|}{Energy($pJ$)} &\multicolumn{2}{c|}{Delay($ns$)}\\ 
\cline{2-5}
&AQFP &CMOS &AQFP &CMOS\\
\hline
4 &5.898E-5 &18.432 &1.2 &614.3 \\
\hline
9 &3.007E-4 &21.504 &2.4 &716.8 \\
\hline
16 &9.063E-4 &23.552 &3.4 &819.2 \\
\hline
25 &1.359E-3 &24.576 &3.6 &819.2 \\
\hline
36 &2.946E-3 &32.768 &5 &921.6 \\
\hline
\end{tabular}
}
\end{table}
The bitonic sorter-based average pooling block using AQPF out-performs its CMOS counterpart by $3.12\times 10^5$ times in energy efficiency and up to 465.46 times in computational speedup. This relative margin is lower because that the CMOS average pooling block implementation simply uses a multiplexer. Therefore the complexity is lower than the proposed bitonic sorter based implementation. However, as aforementioned, the proposed sorter-based average-pooling block can achieve extremely low inaccuracy.   
\begin{table}[b]
\centering
\caption{Hardware utilization of categorization block}
\label{tbl:comp_cat}
\resizebox{\columnwidth}{!}{
\begin{tabular}{|c|c|c|c|c|c|c|}
\hline
\multirow{1}{*}{Input size}&\multicolumn{2}{c|}{Energy($pJ$)} &\multicolumn{2}{c|}{Delay($ns$)}\\ 
\cline{2-5}
&AQFP &CMOS &AQFP &CMOS\\
\hline
100 &1.008E-2 &7825.408 &10 &1945.6 \\
\hline
200 &3.957E-2 &17131.220 &20 &2252.8 \\
\hline
500 &0.244 &37396.480 &50 &2867.2 \\
\hline
800 &0.624 &58880.409 &80 &4300.8 \\
\hline
\end{tabular}
}
\end{table}
Finally, the categorization block can achieve up to $7.76\times 10^5$ times better energy efficiency compared to its CMOS counterpart. Its consumption grows linearly given the majority chain design structure, whose size also grows linearly as input size increases. Overall the hardware consumption of the four basic proposed blocks are far more efficient in energy consumption compared to CMOS-based implementation.
\vspace{-1em}
\subsection{Application Performance}
With all components being well developed and configured, we build a convolutional neural network that has the architecture of Conv3\_x -- AvgPool -- Conv3\_x -- AvgPool -- FC500 -- FC800 -- OutLayer, as indicated in Table \ref{tbl:layer_config} The validity of the network is proven by performing the MNIST \cite{lecun-mnisthandwrittendigit-2010} classification. As the first step to implement a SC-based DNN using AQFP with high inference accuracy, the network is trained with taking all limitations of AQFP and SC into considerations, thereby to prevent the accuracy degradation as much as possible. As shown in Table \ref{tbl:comp_network}, the inference accuracy of this shallow neural network (SNN) is 97.91\% when it is implemented in AQFP, and that of its CMOS counterpart is 97.35\%. Although their accuracy are comparable, the hardware performance of AQFP-based implementation is much more remarkable than the CMOS implementation. The energy improvement of AQFP-based implementation is up to $5.4\times10^4$ times and its throughput improvement is $35.9$ times, comparing to its CMOS counterpart.
\begin{table}[t]
\centering
\caption{DNN Layer Configuration}
\label{tbl:layer_config}
{
\begin{tabular}{|c|c|c|c|}
\hline
Layer Name & Kernel Shape  & Stride  \\
\hline
Conv3\_x   & [$3\times3$, 32]       & 1      \\
\hline
Conv5\_x   & [$5\times5$, 32]       & 1      \\
\hline
Conv7\_x   & [$7\times7$, 64]       & 1       \\
\hline
Conv9\_x   & [$9\times9$, 128]       & 1        \\
\hline
AvgPool    & [$2\times2$]       & 2        \\
\hline
FC500      & 500             & --        \\
\hline
FC800      & 800             & --        \\
\hline
\end{tabular}
}
\end{table}

As discussed above, the AQFP-based neural network can achieve acceptable inference accuracy with remarkable hardware performance when a shallow neural network is implemented. To further explore the feasibility of building a deep neural network in AQFP, we build a deep neural network with the architecture of Conv3\_x -- Conv3\_x -- AvgPool -- Conv5\_x -- Conv5\_x -- AvgPool -- Conv7\_x -- FC500 -- FC800 -- OutLayer, and the configuration of each layer is indicated in Table \ref{tbl:layer_config}. As shown in Table \ref{tbl:comp_network}, the AQFP-based DNN can achieve 96.95\% accuracy in MNIST classification, and the energy efficiency of AQFP is even more remarkable when the network is deeper, the energy improvement is up to $6.9\times10^4$ times and the throughput improvement is $29$ times, comparing to the CMOS-based implementation.
\begin{table}[b]
\centering
\caption{Network Performance Comparison}
\label{tbl:comp_network}
\resizebox{\columnwidth}{!}{
\begin{tabular}{|c|c|c|c|c|c|c|}
\hline
\multicolumn{1}{|l|}{Network} & Platform & Accuracy & Energy($\mu$J) & Throughput(images/ms) \\ \hline
\multirow{3}{*}{SNN}         & Software  &  99.04\%        & --    &   --    \\ \cline{2-5} 
                              & CMOS     &  97.35\%       & 39.46      &  231     \\ \cline{2-5} 
                              & AQFP     &  97.91\%       & 5.606E-4       & 8305      \\ \hline
\multirow{3}{*}{DNN}         & Software  &  99.17\%        &  --      & --      \\ \cline{2-5} 
                              & CMOS     &  96.62\%        &  219.37      &  229     \\ \cline{2-5} 
                              & AQFP     &  96.95\%        & 2.482E-3       & 6667      \\ \hline
\end{tabular}
}
\end{table}

\vspace{-1em}
\section{Conclusion}
In this work, we propose a stochastic-computing deep learning framework using Adiabatic Quantum-Flux-Parametron superconducting technology, which can achieve the highest energy efficiency among superconducting logic families. However, the deep-pipelining nature of AQFP circuits makes the prior design in SC-based DNN not suitable. By taking account of this limitation and other characteristics of AQFP, we redesign the neural network components in SC-based DNN. We propose (i) an accurate integration of summation and activation function using bitonic sorting network and feedback loop, (ii) a low-complexity categorization block based on chain of majority gates, (iii) an ultra-efficient stochastic number generator in AQFP, (iv) a high-accuracy sub-sampling (pooling) block in AQFP, and (v) majority synthesis for further performance improvement and automatic buffer/splitter insertion for requirement of AQFP circuits. Experimental results show that our proposed AQFP-based DNN can achieve up to $6.9\times 10^4$ times higher energy efficiency compared to CMOS-based implementation while maintaining $96\%$ accuracy on the MNIST dataset.

\begin{acks}
This paper is in part supported by National Science Foundation CNS-1704662, CCF-1717754, CCF-1750656, and Defense Advanced Research Projects Agency (DARPA) MTO seedling project.
\end{acks}
\bibliographystyle{ACM-Reference-Format}
\bibliography{ref}


\begin{thebibliography}{55}


\ifx \showCODEN    \undefined \def \showCODEN     #1{\unskip}     \fi
\ifx \showDOI      \undefined \def \showDOI       #1{#1}\fi
\ifx \showISBNx    \undefined \def \showISBNx     #1{\unskip}     \fi
\ifx \showISBNxiii \undefined \def \showISBNxiii  #1{\unskip}     \fi
\ifx \showISSN     \undefined \def \showISSN      #1{\unskip}     \fi
\ifx \showLCCN     \undefined \def \showLCCN      #1{\unskip}     \fi
\ifx \shownote     \undefined \def \shownote      #1{#1}          \fi
\ifx \showarticletitle \undefined \def \showarticletitle #1{#1}   \fi
\ifx \showURL      \undefined \def \showURL       {\relax}        \fi
\providecommand\bibfield[2]{#2}
\providecommand\bibinfo[2]{#2}
\providecommand\natexlab[1]{#1}
\providecommand\showeprint[2][]{arXiv:#2}

\bibitem[\protect\citeauthoryear{Alaghi and Hayes}{Alaghi and Hayes}{2013}]%
        {alaghi2013survey}
\bibfield{author}{\bibinfo{person}{Armin Alaghi} {and} \bibinfo{person}{John~P
  Hayes}.} \bibinfo{year}{2013}\natexlab{}.
\newblock \showarticletitle{Survey of stochastic computing}.
\newblock \bibinfo{journal}{\emph{ACM Transactions on Embedded computing
  systems (TECS)}} \bibinfo{volume}{12}, \bibinfo{number}{2s}
  (\bibinfo{year}{2013}), \bibinfo{pages}{92}.
\newblock


\bibitem[\protect\citeauthoryear{Alaghi, Qian, and Hayes}{Alaghi
  et~al\mbox{.}}{2018}]%
        {alaghi2018promise}
\bibfield{author}{\bibinfo{person}{Armin Alaghi}, \bibinfo{person}{Weikang
  Qian}, {and} \bibinfo{person}{John~P Hayes}.}
  \bibinfo{year}{2018}\natexlab{}.
\newblock \showarticletitle{The promise and challenge of stochastic computing}.
\newblock \bibinfo{journal}{\emph{IEEE Transactions on Computer-Aided Design of
  Integrated Circuits and Systems}} \bibinfo{volume}{37}, \bibinfo{number}{8}
  (\bibinfo{year}{2018}), \bibinfo{pages}{1515--1531}.
\newblock


\bibitem[\protect\citeauthoryear{Ardakani, Leduc-Primeau, Onizawa, Hanyu, and
  Gross}{Ardakani et~al\mbox{.}}{2017}]%
        {ardakani2017vlsi}
\bibfield{author}{\bibinfo{person}{Arash Ardakani},
  \bibinfo{person}{Fran{\c{c}}ois Leduc-Primeau}, \bibinfo{person}{Naoya
  Onizawa}, \bibinfo{person}{Takahiro Hanyu}, {and} \bibinfo{person}{Warren~J
  Gross}.} \bibinfo{year}{2017}\natexlab{}.
\newblock \showarticletitle{VLSI implementation of deep neural network using
  integral stochastic computing}.
\newblock \bibinfo{journal}{\emph{IEEE Transactions on Very Large Scale
  Integration (VLSI) Systems}} \bibinfo{volume}{25}, \bibinfo{number}{10}
  (\bibinfo{year}{2017}), \bibinfo{pages}{2688--2699}.
\newblock


\bibitem[\protect\citeauthoryear{Bang, Wang, Li, Gao, Kim, Dong, Chen, Fick,
  Sun, Dreslinski, Mudge, Kim, Blaauw, and Sylvester}{Bang
  et~al\mbox{.}}{2017}]%
        {bang201714}
\bibfield{author}{\bibinfo{person}{Suyoung Bang}, \bibinfo{person}{Jingcheng
  Wang}, \bibinfo{person}{Ziyun Li}, \bibinfo{person}{Cao Gao},
  \bibinfo{person}{Yejoong Kim}, \bibinfo{person}{Qing Dong},
  \bibinfo{person}{Yen-Po Chen}, \bibinfo{person}{Laura Fick},
  \bibinfo{person}{Xun Sun}, \bibinfo{person}{Ron Dreslinski},
  \bibinfo{person}{Trevor Mudge}, \bibinfo{person}{Hun~Seok Kim},
  \bibinfo{person}{David Blaauw}, {and} \bibinfo{person}{Dennis Sylvester}.}
  \bibinfo{year}{2017}\natexlab{}.
\newblock \showarticletitle{14.7 A 288$\mu$W programmable deep-learning
  processor with 270KB on-chip weight storage using non-uniform memory
  hierarchy for mobile intelligence}. In \bibinfo{booktitle}{\emph{Solid-State
  Circuits Conference (ISSCC), 2017 IEEE International}}. IEEE,
  \bibinfo{pages}{250--251}.
\newblock


\bibitem[\protect\citeauthoryear{Brown and Card}{Brown and Card}{2001}]%
        {brown2001stochastic}
\bibfield{author}{\bibinfo{person}{Bradley~D Brown} {and}
  \bibinfo{person}{Howard~C Card}.} \bibinfo{year}{2001}\natexlab{}.
\newblock \showarticletitle{Stochastic neural computation. I. Computational
  elements}.
\newblock \bibinfo{journal}{\emph{IEEE Transactions on computers}}
  \bibinfo{volume}{50}, \bibinfo{number}{9} (\bibinfo{year}{2001}),
  \bibinfo{pages}{891--905}.
\newblock


\bibitem[\protect\citeauthoryear{Cai, Ren, Liu, Ding, Wang, Qian, Pedram, and
  Wang}{Cai et~al\mbox{.}}{2018}]%
        {Cai:2018:VHA:3296957.3173212}
\bibfield{author}{\bibinfo{person}{Ruizhe Cai}, \bibinfo{person}{Ao Ren},
  \bibinfo{person}{Ning Liu}, \bibinfo{person}{Caiwen Ding},
  \bibinfo{person}{Luhao Wang}, \bibinfo{person}{Xuehai Qian},
  \bibinfo{person}{Massoud Pedram}, {and} \bibinfo{person}{Yanzhi Wang}.}
  \bibinfo{year}{2018}\natexlab{}.
\newblock \showarticletitle{VIBNN: Hardware Acceleration of Bayesian Neural
  Networks}.
\newblock \bibinfo{journal}{\emph{SIGPLAN Not.}} \bibinfo{volume}{53},
  \bibinfo{number}{2} (\bibinfo{date}{March} \bibinfo{year}{2018}),
  \bibinfo{pages}{476--488}.
\newblock
\showISSN{0362-1340}
\urldef\tempurl%
\url{https://doi.org/10.1145/3296957.3173212}
\showDOI{\tempurl}


\bibitem[\protect\citeauthoryear{Chen, Du, Sun, Wang, Wu, Chen, and Temam}{Chen
  et~al\mbox{.}}{2014a}]%
        {chen2014diannao}
\bibfield{author}{\bibinfo{person}{Tianshi Chen}, \bibinfo{person}{Zidong Du},
  \bibinfo{person}{Ninghui Sun}, \bibinfo{person}{Jia Wang},
  \bibinfo{person}{Chengyong Wu}, \bibinfo{person}{Yunji Chen}, {and}
  \bibinfo{person}{Olivier Temam}.} \bibinfo{year}{2014}\natexlab{a}.
\newblock \showarticletitle{Diannao: A small-footprint high-throughput
  accelerator for ubiquitous machine-learning}.
\newblock \bibinfo{journal}{\emph{ACM Sigplan Notices}} \bibinfo{volume}{49},
  \bibinfo{number}{4} (\bibinfo{year}{2014}), \bibinfo{pages}{269--284}.
\newblock


\bibitem[\protect\citeauthoryear{Chen, Luo, Liu, Zhang, He, Wang, Li, Chen, Xu,
  Sun, and Temam}{Chen et~al\mbox{.}}{2014b}]%
        {chen2014dadiannao}
\bibfield{author}{\bibinfo{person}{Yunji Chen}, \bibinfo{person}{Tao Luo},
  \bibinfo{person}{Shaoli Liu}, \bibinfo{person}{Shijin Zhang},
  \bibinfo{person}{Liqiang He}, \bibinfo{person}{Jia Wang},
  \bibinfo{person}{Ling Li}, \bibinfo{person}{Tianshi Chen},
  \bibinfo{person}{Zhiwei Xu}, \bibinfo{person}{Ninghui Sun}, {and}
  \bibinfo{person}{Olivier Temam}.} \bibinfo{year}{2014}\natexlab{b}.
\newblock \showarticletitle{Dadiannao: A machine-learning supercomputer}. In
  \bibinfo{booktitle}{\emph{Proceedings of the 47th Annual IEEE/ACM
  International Symposium on Microarchitecture}}. IEEE Computer Society,
  \bibinfo{pages}{609--622}.
\newblock


\bibitem[\protect\citeauthoryear{Chen, Krishna, Emer, and Sze}{Chen
  et~al\mbox{.}}{2017}]%
        {chen2017eyeriss}
\bibfield{author}{\bibinfo{person}{Yu-Hsin Chen}, \bibinfo{person}{Tushar
  Krishna}, \bibinfo{person}{Joel~S Emer}, {and} \bibinfo{person}{Vivienne
  Sze}.} \bibinfo{year}{2017}\natexlab{}.
\newblock \showarticletitle{Eyeriss: An energy-efficient reconfigurable
  accelerator for deep convolutional neural networks}.
\newblock \bibinfo{journal}{\emph{IEEE Journal of Solid-State Circuits}}
  \bibinfo{volume}{52}, \bibinfo{number}{1} (\bibinfo{year}{2017}),
  \bibinfo{pages}{127--138}.
\newblock


\bibitem[\protect\citeauthoryear{Clarke and Braginski}{Clarke and
  Braginski}{2006}]%
        {clarke2006squid}
\bibfield{author}{\bibinfo{person}{John Clarke} {and} \bibinfo{person}{Alex~I
  Braginski}.} \bibinfo{year}{2006}\natexlab{}.
\newblock \bibinfo{booktitle}{\emph{The SQUID handbook: Applications of SQUIDs
  and SQUID systems}}.
\newblock \bibinfo{publisher}{John Wiley \& Sons}.
\newblock


\bibitem[\protect\citeauthoryear{Courbariaux, Hubara, Soudry, El-Yaniv, and
  Bengio}{Courbariaux et~al\mbox{.}}{2016}]%
        {courbariaux2016binarized}
\bibfield{author}{\bibinfo{person}{Matthieu Courbariaux}, \bibinfo{person}{Itay
  Hubara}, \bibinfo{person}{Daniel Soudry}, \bibinfo{person}{Ran El-Yaniv},
  {and} \bibinfo{person}{Yoshua Bengio}.} \bibinfo{year}{2016}\natexlab{}.
\newblock \showarticletitle{Binarized neural networks: Training deep neural
  networks with weights and activations constrained to+ 1 or-1}.
\newblock \bibinfo{journal}{\emph{arXiv preprint arXiv:1602.02830}}
  (\bibinfo{year}{2016}).
\newblock


\bibitem[\protect\citeauthoryear{Desoli, Chawla, Boesch, Singh, Guidetti,
  De~Ambroggi, Majo, Zambotti, Ayodhyawasi, Singh, and Aggarwal}{Desoli
  et~al\mbox{.}}{2017}]%
        {desoli201714}
\bibfield{author}{\bibinfo{person}{Giuseppe Desoli}, \bibinfo{person}{Nitin
  Chawla}, \bibinfo{person}{Thomas Boesch}, \bibinfo{person}{Surinder-pal
  Singh}, \bibinfo{person}{Elio Guidetti}, \bibinfo{person}{Fabio De~Ambroggi},
  \bibinfo{person}{Tommaso Majo}, \bibinfo{person}{Paolo Zambotti},
  \bibinfo{person}{Manuj Ayodhyawasi}, \bibinfo{person}{Harvinder Singh}, {and}
  \bibinfo{person}{Nalin Aggarwal}.} \bibinfo{year}{2017}\natexlab{}.
\newblock \showarticletitle{14.1 A 2.9 TOPS/W deep convolutional neural network
  SoC in FD-SOI 28nm for intelligent embedded systems}. In
  \bibinfo{booktitle}{\emph{Solid-State Circuits Conference (ISSCC), 2017 IEEE
  International}}. IEEE, \bibinfo{pages}{238--239}.
\newblock


\bibitem[\protect\citeauthoryear{Du, Fasthuber, Chen, Ienne, Li, Luo, Feng,
  Chen, and Temam}{Du et~al\mbox{.}}{2015}]%
        {du2015shidiannao}
\bibfield{author}{\bibinfo{person}{Zidong Du}, \bibinfo{person}{Robert
  Fasthuber}, \bibinfo{person}{Tianshi Chen}, \bibinfo{person}{Paolo Ienne},
  \bibinfo{person}{Ling Li}, \bibinfo{person}{Tao Luo},
  \bibinfo{person}{Xiaobing Feng}, \bibinfo{person}{Yunji Chen}, {and}
  \bibinfo{person}{Olivier Temam}.} \bibinfo{year}{2015}\natexlab{}.
\newblock \showarticletitle{ShiDianNao: Shifting vision processing closer to
  the sensor}. In \bibinfo{booktitle}{\emph{ACM SIGARCH Computer Architecture
  News}}, Vol.~\bibinfo{volume}{43}. ACM, \bibinfo{pages}{92--104}.
\newblock


\bibitem[\protect\citeauthoryear{Gaines}{Gaines}{1967}]%
        {gaines1967stochastic}
\bibfield{author}{\bibinfo{person}{Brian~R Gaines}.}
  \bibinfo{year}{1967}\natexlab{}.
\newblock \showarticletitle{Stochastic computing}. In
  \bibinfo{booktitle}{\emph{Proceedings of the April 18-20, 1967, spring joint
  computer conference}}. ACM, \bibinfo{pages}{149--156}.
\newblock


\bibitem[\protect\citeauthoryear{Gaines}{Gaines}{1969}]%
        {gaines1969stochastic}
\bibfield{author}{\bibinfo{person}{Brian~R Gaines}.}
  \bibinfo{year}{1969}\natexlab{}.
\newblock \showarticletitle{Stochastic computing systems}.
\newblock In \bibinfo{booktitle}{\emph{Advances in information systems
  science}}. \bibinfo{publisher}{Springer}, \bibinfo{pages}{37--172}.
\newblock


\bibitem[\protect\citeauthoryear{Gentle}{Gentle}{2006}]%
        {gentle2006random}
\bibfield{author}{\bibinfo{person}{James~E Gentle}.}
  \bibinfo{year}{2006}\natexlab{}.
\newblock \bibinfo{booktitle}{\emph{Random number generation and Monte Carlo
  methods}}.
\newblock \bibinfo{publisher}{Springer Science \& Business Media}.
\newblock


\bibitem[\protect\citeauthoryear{Han, Kang, Mao, Hu, Li, Li, Xie, Luo, Yao,
  Wang, Yang, and Dally}{Han et~al\mbox{.}}{2017}]%
        {han2017ese}
\bibfield{author}{\bibinfo{person}{Song Han}, \bibinfo{person}{Junlong Kang},
  \bibinfo{person}{Huizi Mao}, \bibinfo{person}{Yiming Hu},
  \bibinfo{person}{Xin Li}, \bibinfo{person}{Yubin Li},
  \bibinfo{person}{Dongliang Xie}, \bibinfo{person}{Hong Luo},
  \bibinfo{person}{Song Yao}, \bibinfo{person}{Yu Wang},
  \bibinfo{person}{Huazhong Yang}, {and} \bibinfo{person}{William~J. Dally}.}
  \bibinfo{year}{2017}\natexlab{}.
\newblock \showarticletitle{ESE: Efficient Speech Recognition Engine with
  Sparse LSTM on FPGA.}. In \bibinfo{booktitle}{\emph{FPGA}}.
  \bibinfo{pages}{75--84}.
\newblock


\bibitem[\protect\citeauthoryear{Han, Liu, Mao, Pu, Pedram, Horowitz, and
  Dally}{Han et~al\mbox{.}}{2016}]%
        {han2016eie}
\bibfield{author}{\bibinfo{person}{Song Han}, \bibinfo{person}{Xingyu Liu},
  \bibinfo{person}{Huizi Mao}, \bibinfo{person}{Jing Pu},
  \bibinfo{person}{Ardavan Pedram}, \bibinfo{person}{Mark~A Horowitz}, {and}
  \bibinfo{person}{William~J Dally}.} \bibinfo{year}{2016}\natexlab{}.
\newblock \showarticletitle{EIE: efficient inference engine on compressed deep
  neural network}. In \bibinfo{booktitle}{\emph{Proceedings of the 43rd
  International Symposium on Computer Architecture}}. IEEE Press,
  \bibinfo{pages}{243--254}.
\newblock


\bibitem[\protect\citeauthoryear{Judd, Albericio, Hetherington, Aamodt, and
  Moshovos}{Judd et~al\mbox{.}}{2016}]%
        {judd2016stripes}
\bibfield{author}{\bibinfo{person}{Patrick Judd}, \bibinfo{person}{Jorge
  Albericio}, \bibinfo{person}{Tayler Hetherington}, \bibinfo{person}{Tor~M
  Aamodt}, {and} \bibinfo{person}{Andreas Moshovos}.}
  \bibinfo{year}{2016}\natexlab{}.
\newblock \showarticletitle{Stripes: Bit-serial deep neural network computing}.
  In \bibinfo{booktitle}{\emph{Microarchitecture (MICRO), 2016 49th Annual
  IEEE/ACM International Symposium on}}. IEEE, \bibinfo{pages}{1--12}.
\newblock


\bibitem[\protect\citeauthoryear{Kwon, Samajdar, and Krishna}{Kwon
  et~al\mbox{.}}{2018}]%
        {kwon2018maeri}
\bibfield{author}{\bibinfo{person}{Hyoukjun Kwon}, \bibinfo{person}{Ananda
  Samajdar}, {and} \bibinfo{person}{Tushar Krishna}.}
  \bibinfo{year}{2018}\natexlab{}.
\newblock \showarticletitle{MAERI: Enabling Flexible Dataflow Mapping over DNN
  Accelerators via Reconfigurable Interconnects}. In
  \bibinfo{booktitle}{\emph{Proceedings of the Twenty-Third International
  Conference on Architectural Support for Programming Languages and Operating
  Systems}}. ACM, \bibinfo{pages}{461--475}.
\newblock


\bibitem[\protect\citeauthoryear{LeCun and Cortes}{LeCun and Cortes}{2010}]%
        {lecun-mnisthandwrittendigit-2010}
\bibfield{author}{\bibinfo{person}{Yann LeCun} {and} \bibinfo{person}{Corinna
  Cortes}.} \bibinfo{year}{2010}\natexlab{}.
\newblock \showarticletitle{{MNIST} handwritten digit database}.
\newblock \bibinfo{howpublished}{http://yann.lecun.com/exdb/mnist/}.
\newblock  (\bibinfo{year}{2010}).
\newblock
\urldef\tempurl%
\url{http://yann.lecun.com/exdb/mnist/}
\showURL{%
\tempurl}


\bibitem[\protect\citeauthoryear{Lee, Alaghi, Hayes, Sathe, and Ceze}{Lee
  et~al\mbox{.}}{2017}]%
        {lee2017energy}
\bibfield{author}{\bibinfo{person}{Vincent~T Lee}, \bibinfo{person}{Armin
  Alaghi}, \bibinfo{person}{John~P Hayes}, \bibinfo{person}{Visvesh Sathe},
  {and} \bibinfo{person}{Luis Ceze}.} \bibinfo{year}{2017}\natexlab{}.
\newblock \showarticletitle{Energy-efficient hybrid stochastic-binary neural
  networks for near-sensor computing}. In \bibinfo{booktitle}{\emph{Proceedings
  of the Conference on Design, Automation \& Test in Europe}}. European Design
  and Automation Association, \bibinfo{pages}{13--18}.
\newblock


\bibitem[\protect\citeauthoryear{{Likharev}}{{Likharev}}{1977}]%
        {Likharev}
\bibfield{author}{\bibinfo{person}{K. {Likharev}}.}
  \bibinfo{year}{1977}\natexlab{}.
\newblock \showarticletitle{Dynamics of some single flux quantum devices: I.
  Parametric quantron}.
\newblock \bibinfo{journal}{\emph{IEEE Transactions on Magnetics}}
  \bibinfo{volume}{13}, \bibinfo{number}{1} (\bibinfo{date}{January}
  \bibinfo{year}{1977}), \bibinfo{pages}{242--244}.
\newblock
\showISSN{0018-9464}
\urldef\tempurl%
\url{https://doi.org/10.1109/TMAG.1977.1059351}
\showDOI{\tempurl}


\bibitem[\protect\citeauthoryear{Likharev and Semenov}{Likharev and
  Semenov}{1991}]%
        {80745}
\bibfield{author}{\bibinfo{person}{K.~K. Likharev} {and} \bibinfo{person}{V.~K.
  Semenov}.} \bibinfo{year}{1991}\natexlab{}.
\newblock \showarticletitle{RSFQ logic/memory family: a new Josephson-junction
  technology for sub-terahertz-clock-frequency digital systems}.
\newblock \bibinfo{journal}{\emph{IEEE Transactions on Applied
  Superconductivity}} \bibinfo{volume}{1}, \bibinfo{number}{1}
  (\bibinfo{date}{March} \bibinfo{year}{1991}), \bibinfo{pages}{3--28}.
\newblock
\showISSN{1051-8223}
\urldef\tempurl%
\url{https://doi.org/10.1109/77.80745}
\showDOI{\tempurl}


\bibitem[\protect\citeauthoryear{Liszka and Batcher}{Liszka and
  Batcher}{1993}]%
        {liszka1993generalized}
\bibfield{author}{\bibinfo{person}{Kathy~J Liszka} {and}
  \bibinfo{person}{Kenneth~E Batcher}.} \bibinfo{year}{1993}\natexlab{}.
\newblock \showarticletitle{A generalized bitonic sorting network}. In
  \bibinfo{booktitle}{\emph{Parallel Processing, 1993. ICPP 1993. International
  Conference on}}, Vol.~\bibinfo{volume}{1}. IEEE, \bibinfo{pages}{105--108}.
\newblock


\bibitem[\protect\citeauthoryear{{Loe} and {Goto}}{{Loe} and {Goto}}{1985}]%
        {Goto}
\bibfield{author}{\bibinfo{person}{K. {Loe}} {and} \bibinfo{person}{E.
  {Goto}}.} \bibinfo{year}{1985}\natexlab{}.
\newblock \showarticletitle{Analysis of flux input and output Josephson pair
  device}.
\newblock \bibinfo{journal}{\emph{IEEE Transactions on Magnetics}}
  \bibinfo{volume}{21}, \bibinfo{number}{2} (\bibinfo{date}{March}
  \bibinfo{year}{1985}), \bibinfo{pages}{884--887}.
\newblock
\showISSN{0018-9464}
\urldef\tempurl%
\url{https://doi.org/10.1109/TMAG.1985.1063734}
\showDOI{\tempurl}


\bibitem[\protect\citeauthoryear{L’Ecuyer}{L’Ecuyer}{2012}]%
        {l2012random}
\bibfield{author}{\bibinfo{person}{Pierre L’Ecuyer}.}
  \bibinfo{year}{2012}\natexlab{}.
\newblock \showarticletitle{Random number generation}.
\newblock In \bibinfo{booktitle}{\emph{Handbook of Computational Statistics}}.
  \bibinfo{publisher}{Springer}, \bibinfo{pages}{35--71}.
\newblock


\bibitem[\protect\citeauthoryear{Mahajan, Park, Amaro, Sharma, Yazdanbakhsh,
  Kim, and Esmaeilzadeh}{Mahajan et~al\mbox{.}}{2016}]%
        {mahajan2016tabla}
\bibfield{author}{\bibinfo{person}{Divya Mahajan}, \bibinfo{person}{Jongse
  Park}, \bibinfo{person}{Emmanuel Amaro}, \bibinfo{person}{Hardik Sharma},
  \bibinfo{person}{Amir Yazdanbakhsh}, \bibinfo{person}{Joon~Kyung Kim}, {and}
  \bibinfo{person}{Hadi Esmaeilzadeh}.} \bibinfo{year}{2016}\natexlab{}.
\newblock \showarticletitle{Tabla: A unified template-based framework for
  accelerating statistical machine learning}. In \bibinfo{booktitle}{\emph{High
  Performance Computer Architecture (HPCA), 2016 IEEE International Symposium
  on}}. IEEE, \bibinfo{pages}{14--26}.
\newblock


\bibitem[\protect\citeauthoryear{Moons, Uytterhoeven, Dehaene, and
  Verhelst}{Moons et~al\mbox{.}}{2017}]%
        {moons201714}
\bibfield{author}{\bibinfo{person}{Bert Moons}, \bibinfo{person}{Roel
  Uytterhoeven}, \bibinfo{person}{Wim Dehaene}, {and} \bibinfo{person}{Marian
  Verhelst}.} \bibinfo{year}{2017}\natexlab{}.
\newblock \showarticletitle{14.5 Envision: A 0.26-to-10TOPS/W subword-parallel
  dynamic-voltage-accuracy-frequency-scalable Convolutional Neural Network
  processor in 28nm FDSOI}. In \bibinfo{booktitle}{\emph{Solid-State Circuits
  Conference (ISSCC), 2017 IEEE International}}. IEEE,
  \bibinfo{pages}{246--247}.
\newblock


\bibitem[\protect\citeauthoryear{Nagasawa, Hashimoto, Numata, and
  Tahara}{Nagasawa et~al\mbox{.}}{1995}]%
        {nagasawa1995380}
\bibfield{author}{\bibinfo{person}{Shuichi Nagasawa},
  \bibinfo{person}{Yoshihito Hashimoto}, \bibinfo{person}{Hideaki Numata},
  {and} \bibinfo{person}{Shuichi Tahara}.} \bibinfo{year}{1995}\natexlab{}.
\newblock \showarticletitle{A 380 ps, 9.5 mW Josephson 4-Kbit RAM operated at a
  high bit yield}.
\newblock \bibinfo{journal}{\emph{IEEE Transactions on Applied
  Superconductivity}} \bibinfo{volume}{5}, \bibinfo{number}{2}
  (\bibinfo{year}{1995}), \bibinfo{pages}{2447--2452}.
\newblock


\bibitem[\protect\citeauthoryear{Narama, China, Takeuchi, Ortlepp, Yamanashi,
  and Yoshikawa}{Narama et~al\mbox{.}}{2016}]%
        {83k}
\bibfield{author}{\bibinfo{person}{T. Narama}, \bibinfo{person}{F. China},
  \bibinfo{person}{N. Takeuchi}, \bibinfo{person}{T. Ortlepp},
  \bibinfo{person}{Y. Yamanashi}, {and} \bibinfo{person}{N. Yoshikawa}.}
  \bibinfo{year}{2016}\natexlab{}.
\newblock \showarticletitle{Yield evaluation of 83k-junction
  adiabatic-quantum-flux-parametron circuit}. In \bibinfo{booktitle}{\emph{2016
  Appl. Superconductivity Conference (ASC2016)}}.
\newblock


\bibitem[\protect\citeauthoryear{Niederreiter}{Niederreiter}{1992}]%
        {niederreiter1992random}
\bibfield{author}{\bibinfo{person}{Harald Niederreiter}.}
  \bibinfo{year}{1992}\natexlab{}.
\newblock \bibinfo{booktitle}{\emph{Random number generation and quasi-Monte
  Carlo methods}}. Vol.~\bibinfo{volume}{63}.
\newblock \bibinfo{publisher}{Siam}.
\newblock


\bibitem[\protect\citeauthoryear{Qiu, Wang, Yao, Guo, Li, Zhou, Yu, Tang, Xu,
  Song, Wang, and Yang}{Qiu et~al\mbox{.}}{2016}]%
        {qiu2016going}
\bibfield{author}{\bibinfo{person}{Jiantao Qiu}, \bibinfo{person}{Jie Wang},
  \bibinfo{person}{Song Yao}, \bibinfo{person}{Kaiyuan Guo},
  \bibinfo{person}{Boxun Li}, \bibinfo{person}{Erjin Zhou},
  \bibinfo{person}{Jincheng Yu}, \bibinfo{person}{Tianqi Tang},
  \bibinfo{person}{Ningyi Xu}, \bibinfo{person}{Sen Song}, \bibinfo{person}{Yu
  Wang}, {and} \bibinfo{person}{Huazhong Yang}.}
  \bibinfo{year}{2016}\natexlab{}.
\newblock \showarticletitle{Going deeper with embedded fpga platform for
  convolutional neural network}. In \bibinfo{booktitle}{\emph{Proceedings of
  the 2016 ACM/SIGDA International Symposium on Field-Programmable Gate
  Arrays}}. ACM, \bibinfo{pages}{26--35}.
\newblock


\bibitem[\protect\citeauthoryear{Reagen, Whatmough, Adolf, Rama, Lee, Lee,
  Hern{\'a}ndez-Lobato, Wei, and Brooks}{Reagen et~al\mbox{.}}{2016}]%
        {reagen2016minerva}
\bibfield{author}{\bibinfo{person}{Brandon Reagen}, \bibinfo{person}{Paul
  Whatmough}, \bibinfo{person}{Robert Adolf}, \bibinfo{person}{Saketh Rama},
  \bibinfo{person}{Hyunkwang Lee}, \bibinfo{person}{Sae~Kyu Lee},
  \bibinfo{person}{Jos{\'e}~Miguel Hern{\'a}ndez-Lobato},
  \bibinfo{person}{Gu-Yeon Wei}, {and} \bibinfo{person}{David Brooks}.}
  \bibinfo{year}{2016}\natexlab{}.
\newblock \showarticletitle{Minerva: Enabling low-power, highly-accurate deep
  neural network accelerators}. In \bibinfo{booktitle}{\emph{Proceedings of the
  43rd International Symposium on Computer Architecture}}. IEEE Press,
  \bibinfo{pages}{267--278}.
\newblock


\bibitem[\protect\citeauthoryear{Ren, Li, Ding, Qiu, Wang, Li, Qian, and
  Yuan}{Ren et~al\mbox{.}}{2017}]%
        {ren2017sc}
\bibfield{author}{\bibinfo{person}{Ao Ren}, \bibinfo{person}{Zhe Li},
  \bibinfo{person}{Caiwen Ding}, \bibinfo{person}{Qinru Qiu},
  \bibinfo{person}{Yanzhi Wang}, \bibinfo{person}{Ji Li},
  \bibinfo{person}{Xuehai Qian}, {and} \bibinfo{person}{Bo Yuan}.}
  \bibinfo{year}{2017}\natexlab{}.
\newblock \showarticletitle{Sc-dcnn: Highly-scalable deep convolutional neural
  network using stochastic computing}.
\newblock \bibinfo{journal}{\emph{ACM SIGOPS Operating Systems Review}}
  \bibinfo{volume}{51}, \bibinfo{number}{2} (\bibinfo{year}{2017}),
  \bibinfo{pages}{405--418}.
\newblock


\bibitem[\protect\citeauthoryear{Ren, Li, Wang, Qiu, and Yuan}{Ren
  et~al\mbox{.}}{2016}]%
        {ren2016designing}
\bibfield{author}{\bibinfo{person}{Ao Ren}, \bibinfo{person}{Zhe Li},
  \bibinfo{person}{Yanzhi Wang}, \bibinfo{person}{Qinru Qiu}, {and}
  \bibinfo{person}{Bo Yuan}.} \bibinfo{year}{2016}\natexlab{}.
\newblock \showarticletitle{Designing reconfigurable large-scale deep learning
  systems using stochastic computing}. In \bibinfo{booktitle}{\emph{Rebooting
  Computing (ICRC), IEEE International Conference on}}. IEEE,
  \bibinfo{pages}{1--7}.
\newblock


\bibitem[\protect\citeauthoryear{Sharma, Park, Mahajan, Amaro, Kim, Shao,
  Mishra, and Esmaeilzadeh}{Sharma et~al\mbox{.}}{2016}]%
        {sharma2016high}
\bibfield{author}{\bibinfo{person}{Hardik Sharma}, \bibinfo{person}{Jongse
  Park}, \bibinfo{person}{Divya Mahajan}, \bibinfo{person}{Emmanuel Amaro},
  \bibinfo{person}{Joon~Kyung Kim}, \bibinfo{person}{Chenkai Shao},
  \bibinfo{person}{Asit Mishra}, {and} \bibinfo{person}{Hadi Esmaeilzadeh}.}
  \bibinfo{year}{2016}\natexlab{}.
\newblock \showarticletitle{From high-level deep neural models to FPGAs}. In
  \bibinfo{booktitle}{\emph{Microarchitecture (MICRO), 2016 49th Annual
  IEEE/ACM International Symposium on}}. IEEE, \bibinfo{pages}{1--12}.
\newblock


\bibitem[\protect\citeauthoryear{Sim and Lee}{Sim and Lee}{2017}]%
        {sim2017new}
\bibfield{author}{\bibinfo{person}{Hyeonuk Sim} {and} \bibinfo{person}{Jongeun
  Lee}.} \bibinfo{year}{2017}\natexlab{}.
\newblock \showarticletitle{A new stochastic computing multiplier with
  application to deep convolutional neural networks}. In
  \bibinfo{booktitle}{\emph{Proceedings of the 54th Annual Design Automation
  Conference 2017}}. ACM, \bibinfo{pages}{29}.
\newblock


\bibitem[\protect\citeauthoryear{Sim, Park, Kim, Bae, Choi, and Kim}{Sim
  et~al\mbox{.}}{2016}]%
        {sim201614}
\bibfield{author}{\bibinfo{person}{Jaehyeong Sim}, \bibinfo{person}{Jun-Seok
  Park}, \bibinfo{person}{Minhye Kim}, \bibinfo{person}{Dongmyung Bae},
  \bibinfo{person}{Yeongjae Choi}, {and} \bibinfo{person}{Lee-Sup Kim}.}
  \bibinfo{year}{2016}\natexlab{}.
\newblock \showarticletitle{14.6 a 1.42 tops/w deep convolutional neural
  network recognition processor for intelligent ioe systems}. In
  \bibinfo{booktitle}{\emph{Solid-State Circuits Conference (ISSCC), 2016 IEEE
  International}}. IEEE, \bibinfo{pages}{264--265}.
\newblock


\bibitem[\protect\citeauthoryear{Song, Zhong, Zhang, Hu, Liu, Zhang, Wang, and
  Li}{Song et~al\mbox{.}}{2018}]%
        {song2018situ}
\bibfield{author}{\bibinfo{person}{Mingcong Song}, \bibinfo{person}{Kan Zhong},
  \bibinfo{person}{Jiaqi Zhang}, \bibinfo{person}{Yang Hu},
  \bibinfo{person}{Duo Liu}, \bibinfo{person}{Weigong Zhang},
  \bibinfo{person}{Jing Wang}, {and} \bibinfo{person}{Tao Li}.}
  \bibinfo{year}{2018}\natexlab{}.
\newblock \showarticletitle{In-Situ AI: Towards Autonomous and Incremental Deep
  Learning for IoT Systems}. In \bibinfo{booktitle}{\emph{High Performance
  Computer Architecture (HPCA), 2018 IEEE International Symposium on}}. IEEE,
  \bibinfo{pages}{92--103}.
\newblock


\bibitem[\protect\citeauthoryear{Suda, Chandra, Dasika, Mohanty, Ma, Vrudhula,
  Seo, and Cao}{Suda et~al\mbox{.}}{2016}]%
        {suda2016throughput}
\bibfield{author}{\bibinfo{person}{Naveen Suda}, \bibinfo{person}{Vikas
  Chandra}, \bibinfo{person}{Ganesh Dasika}, \bibinfo{person}{Abinash Mohanty},
  \bibinfo{person}{Yufei Ma}, \bibinfo{person}{Sarma Vrudhula},
  \bibinfo{person}{Jae-sun Seo}, {and} \bibinfo{person}{Yu Cao}.}
  \bibinfo{year}{2016}\natexlab{}.
\newblock \showarticletitle{Throughput-optimized OpenCL-based FPGA accelerator
  for large-scale convolutional neural networks}. In
  \bibinfo{booktitle}{\emph{Proceedings of the 2016 ACM/SIGDA International
  Symposium on Field-Programmable Gate Arrays}}. ACM, \bibinfo{pages}{16--25}.
\newblock


\bibitem[\protect\citeauthoryear{Takeuchi, Nagasawa, China, Ando, Hidaka,
  Yamanashi, and Yoshikawa}{Takeuchi et~al\mbox{.}}{2017}]%
        {takeuchi2017adiabatic}
\bibfield{author}{\bibinfo{person}{Naoki Takeuchi}, \bibinfo{person}{Shuichi
  Nagasawa}, \bibinfo{person}{Fumihiro China}, \bibinfo{person}{Takumi Ando},
  \bibinfo{person}{Mutsuo Hidaka}, \bibinfo{person}{Yuki Yamanashi}, {and}
  \bibinfo{person}{Nobuyuki Yoshikawa}.} \bibinfo{year}{2017}\natexlab{}.
\newblock \showarticletitle{Adiabatic quantum-flux-parametron cell library
  designed using a 10 kA cm- 2 niobium fabrication process}.
\newblock \bibinfo{journal}{\emph{Superconductor Science and Technology}}
  \bibinfo{volume}{30}, \bibinfo{number}{3} (\bibinfo{year}{2017}),
  \bibinfo{pages}{035002}.
\newblock


\bibitem[\protect\citeauthoryear{Takeuchi, Ozawa, Yamanashi, and
  Yoshikawa}{Takeuchi et~al\mbox{.}}{2013a}]%
        {takeuchi2013adiabatic}
\bibfield{author}{\bibinfo{person}{Naoki Takeuchi}, \bibinfo{person}{Dan
  Ozawa}, \bibinfo{person}{Yuki Yamanashi}, {and} \bibinfo{person}{Nobuyuki
  Yoshikawa}.} \bibinfo{year}{2013}\natexlab{a}.
\newblock \showarticletitle{An adiabatic quantum flux parametron as an
  ultra-low-power logic device}.
\newblock \bibinfo{journal}{\emph{Superconductor Science and Technology}}
  \bibinfo{volume}{26}, \bibinfo{number}{3} (\bibinfo{year}{2013}),
  \bibinfo{pages}{035010}.
\newblock


\bibitem[\protect\citeauthoryear{Takeuchi, Yamanashi, and Yoshikawa}{Takeuchi
  et~al\mbox{.}}{2013b}]%
        {takeuchi2013measurement}
\bibfield{author}{\bibinfo{person}{Naoki Takeuchi}, \bibinfo{person}{Yuki
  Yamanashi}, {and} \bibinfo{person}{Nobuyuki Yoshikawa}.}
  \bibinfo{year}{2013}\natexlab{b}.
\newblock \showarticletitle{Measurement of 10 zJ energy dissipation of
  adiabatic quantum-flux-parametron logic using a superconducting resonator}.
\newblock \bibinfo{journal}{\emph{Applied Physics Letters}}
  \bibinfo{volume}{102}, \bibinfo{number}{5} (\bibinfo{year}{2013}),
  \bibinfo{pages}{052602}.
\newblock


\bibitem[\protect\citeauthoryear{Takeuchi, Yamanashi, and Yoshikawa}{Takeuchi
  et~al\mbox{.}}{2014}]%
        {Takeuchi2014QuantumLimits}
\bibfield{author}{\bibinfo{person}{Naoki Takeuchi}, \bibinfo{person}{Yuki
  Yamanashi}, {and} \bibinfo{person}{Nobuyuki Yoshikawa}.}
  \bibinfo{year}{2014}\natexlab{}.
\newblock \showarticletitle{Energy efficiency of adiabatic superconductor
  logic}.
\newblock \bibinfo{journal}{\emph{Superconductor Science and Technology}}
  \bibinfo{volume}{28}, \bibinfo{number}{1} (\bibinfo{date}{nov}
  \bibinfo{year}{2014}), \bibinfo{pages}{015003}.
\newblock
\urldef\tempurl%
\url{https://doi.org/10.1088/0953-2048/28/1/015003}
\showDOI{\tempurl}


\bibitem[\protect\citeauthoryear{Takeuchi, Yamanashi, and Yoshikawa}{Takeuchi
  et~al\mbox{.}}{2015}]%
        {takeuchi2015adiabatic}
\bibfield{author}{\bibinfo{person}{Naoki Takeuchi}, \bibinfo{person}{Yuki
  Yamanashi}, {and} \bibinfo{person}{Nobuyuki Yoshikawa}.}
  \bibinfo{year}{2015}\natexlab{}.
\newblock \showarticletitle{Adiabatic quantum-flux-parametron cell library
  adopting minimalist design}.
\newblock \bibinfo{journal}{\emph{Journal of Applied Physics}}
  \bibinfo{volume}{117}, \bibinfo{number}{17} (\bibinfo{year}{2015}),
  \bibinfo{pages}{173912}.
\newblock


\bibitem[\protect\citeauthoryear{Tolpygo, Bolkhovsky, Weir, Wynn, Oates,
  Johnson, and Gouker}{Tolpygo et~al\mbox{.}}{2016}]%
        {tolpygo2016advanced}
\bibfield{author}{\bibinfo{person}{Sergey~K Tolpygo}, \bibinfo{person}{Vladimir
  Bolkhovsky}, \bibinfo{person}{Terence~J Weir}, \bibinfo{person}{Alex Wynn},
  \bibinfo{person}{Daniel~E Oates}, \bibinfo{person}{Leonard~M Johnson}, {and}
  \bibinfo{person}{Mark~A Gouker}.} \bibinfo{year}{2016}\natexlab{}.
\newblock \showarticletitle{Advanced fabrication processes for superconducting
  very large-scale integrated circuits}.
\newblock \bibinfo{journal}{\emph{IEEE Transactions on Applied
  Superconductivity}} \bibinfo{volume}{26}, \bibinfo{number}{3}
  (\bibinfo{year}{2016}), \bibinfo{pages}{1--10}.
\newblock


\bibitem[\protect\citeauthoryear{Umuroglu, Fraser, Gambardella, Blott, Leong,
  Jahre, and Vissers}{Umuroglu et~al\mbox{.}}{2017}]%
        {umuroglu2017finn}
\bibfield{author}{\bibinfo{person}{Yaman Umuroglu}, \bibinfo{person}{Nicholas~J
  Fraser}, \bibinfo{person}{Giulio Gambardella}, \bibinfo{person}{Michaela
  Blott}, \bibinfo{person}{Philip Leong}, \bibinfo{person}{Magnus Jahre}, {and}
  \bibinfo{person}{Kees Vissers}.} \bibinfo{year}{2017}\natexlab{}.
\newblock \showarticletitle{Finn: A framework for fast, scalable binarized
  neural network inference}. In \bibinfo{booktitle}{\emph{Proceedings of the
  2017 ACM/SIGDA International Symposium on Field-Programmable Gate Arrays}}.
  ACM, \bibinfo{pages}{65--74}.
\newblock


\bibitem[\protect\citeauthoryear{Venkataramani, Ranjan, Banerjee, Das, Avancha,
  Jagannathan, Durg, Nagaraj, Kaul, Dubey, and Raghunathan}{Venkataramani
  et~al\mbox{.}}{2017}]%
        {venkataramani2017scaledeep}
\bibfield{author}{\bibinfo{person}{Swagath Venkataramani},
  \bibinfo{person}{Ashish Ranjan}, \bibinfo{person}{Subarno Banerjee},
  \bibinfo{person}{Dipankar Das}, \bibinfo{person}{Sasikanth Avancha},
  \bibinfo{person}{Ashok Jagannathan}, \bibinfo{person}{Ajaya Durg},
  \bibinfo{person}{Dheemanth Nagaraj}, \bibinfo{person}{Bharat Kaul},
  \bibinfo{person}{Pradeep Dubey}, {and} \bibinfo{person}{Anand Raghunathan}.}
  \bibinfo{year}{2017}\natexlab{}.
\newblock \showarticletitle{ScaleDeep: A Scalable Compute Architecture for
  Learning and Evaluating Deep Networks}. In
  \bibinfo{booktitle}{\emph{Proceedings of the 44th Annual International
  Symposium on Computer Architecture}}. ACM, \bibinfo{pages}{13--26}.
\newblock


\bibitem[\protect\citeauthoryear{Wang, Zhan, Li, Tang, Yuan, Zhao, Wen, Wang,
  and Lin}{Wang et~al\mbox{.}}{2018}]%
        {wang2018universal}
\bibfield{author}{\bibinfo{person}{Yanzhi Wang}, \bibinfo{person}{Zheng Zhan},
  \bibinfo{person}{Jiayu Li}, \bibinfo{person}{Jian Tang}, \bibinfo{person}{Bo
  Yuan}, \bibinfo{person}{Liang Zhao}, \bibinfo{person}{Wujie Wen},
  \bibinfo{person}{Siyue Wang}, {and} \bibinfo{person}{Xue Lin}.}
  \bibinfo{year}{2018}\natexlab{}.
\newblock \showarticletitle{On the Universal Approximation Property and
  Equivalence of Stochastic Computing-based Neural Networks and Binary Neural
  Networks}.
\newblock \bibinfo{journal}{\emph{arXiv preprint arXiv:1803.05391}}
  (\bibinfo{year}{2018}).
\newblock


\bibitem[\protect\citeauthoryear{Whatmough, Lee, Lee, Rama, Brooks, and
  Wei}{Whatmough et~al\mbox{.}}{2017}]%
        {whatmough201714}
\bibfield{author}{\bibinfo{person}{Paul~N Whatmough}, \bibinfo{person}{Sae~Kyu
  Lee}, \bibinfo{person}{Hyunkwang Lee}, \bibinfo{person}{Saketh Rama},
  \bibinfo{person}{David Brooks}, {and} \bibinfo{person}{Gu-Yeon Wei}.}
  \bibinfo{year}{2017}\natexlab{}.
\newblock \showarticletitle{14.3 A 28nm SoC with a 1.2 GHz 568nJ/prediction
  sparse deep-neural-network engine with> 0.1 timing error rate tolerance for
  IoT applications}. In \bibinfo{booktitle}{\emph{Solid-State Circuits
  Conference (ISSCC), 2017 IEEE International}}. IEEE,
  \bibinfo{pages}{242--243}.
\newblock


\bibitem[\protect\citeauthoryear{Xiu}{Xiu}{2010}]%
        {xiu2010numerical}
\bibfield{author}{\bibinfo{person}{Dongbin Xiu}.}
  \bibinfo{year}{2010}\natexlab{}.
\newblock \bibinfo{booktitle}{\emph{Numerical methods for stochastic
  computations: a spectral method approach}}.
\newblock \bibinfo{publisher}{Princeton university press}.
\newblock


\bibitem[\protect\citeauthoryear{Zhang, Fang, Zhou, Pan, and Cong}{Zhang
  et~al\mbox{.}}{2016a}]%
        {zhang2016caffeine}
\bibfield{author}{\bibinfo{person}{Chen Zhang}, \bibinfo{person}{Zhenman Fang},
  \bibinfo{person}{Peipei Zhou}, \bibinfo{person}{Peichen Pan}, {and}
  \bibinfo{person}{Jason Cong}.} \bibinfo{year}{2016}\natexlab{a}.
\newblock \showarticletitle{Caffeine: Towards uniformed representation and
  acceleration for deep convolutional neural networks}. In
  \bibinfo{booktitle}{\emph{Computer-Aided Design (ICCAD), 2016 IEEE/ACM
  International Conference on}}. IEEE, \bibinfo{pages}{1--8}.
\newblock


\bibitem[\protect\citeauthoryear{Zhang, Wu, Sun, Sun, Luo, and Cong}{Zhang
  et~al\mbox{.}}{2016b}]%
        {zhang2016energy}
\bibfield{author}{\bibinfo{person}{Chen Zhang}, \bibinfo{person}{Di Wu},
  \bibinfo{person}{Jiayu Sun}, \bibinfo{person}{Guangyu Sun},
  \bibinfo{person}{Guojie Luo}, {and} \bibinfo{person}{Jason Cong}.}
  \bibinfo{year}{2016}\natexlab{b}.
\newblock \showarticletitle{Energy-efficient CNN implementation on a deeply
  pipelined FPGA cluster}. In \bibinfo{booktitle}{\emph{Proceedings of the 2016
  International Symposium on Low Power Electronics and Design}}. ACM,
  \bibinfo{pages}{326--331}.
\newblock


\bibitem[\protect\citeauthoryear{Zhao, Song, Zhang, Xing, Lin, Srivastava,
  Gupta, and Zhang}{Zhao et~al\mbox{.}}{2017}]%
        {zhao2017accelerating}
\bibfield{author}{\bibinfo{person}{Ritchie Zhao}, \bibinfo{person}{Weinan
  Song}, \bibinfo{person}{Wentao Zhang}, \bibinfo{person}{Tianwei Xing},
  \bibinfo{person}{Jeng-Hau Lin}, \bibinfo{person}{Mani~B Srivastava},
  \bibinfo{person}{Rajesh Gupta}, {and} \bibinfo{person}{Zhiru Zhang}.}
  \bibinfo{year}{2017}\natexlab{}.
\newblock \showarticletitle{Accelerating Binarized Convolutional Neural
  Networks with Software-Programmable FPGAs.}. In
  \bibinfo{booktitle}{\emph{FPGA}}. \bibinfo{pages}{15--24}.
\newblock


\end{thebibliography}

\end{document}